\documentclass[12pt,english,a4paper]{article}

\renewcommand{\footnotemark}{}

\usepackage{enumitem}
\usepackage{babel}
\usepackage[T1]{fontenc}
\usepackage[latin1]{inputenc}
\usepackage{graphicx}
\usepackage{amssymb}
\usepackage{amsmath}

\begin{document}

\title{\bf \LARGE Transparallel mind:\\Classical computing with quantum power}

\author{{\Large Peter A. van der Helm}\\[5mm]
{\it Laboratory of Experimental Psychology}\\
{\it University of Leuven (K.U. Leuven)}\\
{\it Tiensestraat 102 - box 3711, Leuven B-3000, Belgium}\\[5mm]
{\it E: peter.vanderhelm@ppw.kuleuven.be}\\
{\it U: https://perswww.kuleuven.be/peter\_van\_der\_helm}}

\date{~}

\maketitle

\newpage{}

\section*{Abstract}

Inspired by the extraordinary computing power promised by quantum computers,
the quantum mind hypothesis postulated that quantum mechanical phenomena are
the source of neuronal synchronization, which, in turn, might underlie
consciousness.
Here, I present an alternative inspired by a classical computing method with
quantum power.
This method relies on special distributed representations called hyperstrings.
Hyperstrings are superpositions of up to an exponential number of
strings, which -- by a single-processor classical computer -- can be evaluated
in a transparallel fashion, that is, simultaneously as if only one string
were concerned.
Building on a neurally plausible model of human visual perceptual
organization, in which hyperstrings are formal counterparts of transient
neural assemblies, I postulate that synchronization in such assemblies is a
manifestation of transparallel information processing.
This accounts for the high combinatorial capacity and speed of human visual
perceptual organization and strengthens ideas that self-organizing cognitive
architecture bridges the gap between neurons and consciousness.


\section*{Keywords}

cognitive architecture;
distributed representations;
neuronal synchronization;
quantum computing;
transparallel computing by hyperstrings;
visual perceptual organization.


\section*{Highlights}

\begin{itemize}

\item[$\bullet$]
A neurally plausible alternative to the quantum mind hypothesis is presented.

\item[$\bullet$]
This alternative is based on a classical computing method with quantum power.

\item[$\bullet$]
This method relies on special distributed representations, called
hyperstrings.

\item[$\bullet$]
Hyperstrings allow many similar features to be processed as one feature.

\item[$\bullet$]
Thereby, they enable a computational explanation of neuronal synchronization.

\end{itemize}

\newpage{}

\section{Introduction}

Mind usually is taken to refer to cognitive faculties, such as perception and
memory, which enable consciousness.
In this article at the intersection of cognitive science and artificial
intelligence research, I do not discuss full-blown models of cognition as a
whole, but I do aim to shed more light on the nature of cognitive processes.
To this end, I review a powerful classical computing method -- called
transparallel processing by hyperstrings (van der Helm 2004) -- which has been
implemented in a minimal coding algorithm called PISA.\footnote{\label{pisa}
PISA is available at
https://perswww.kuleuven.be/$\sim$u0084530/doc/pisa.html.
Its worst-case computing time may be weakly-exponential (i.e.,
near-tractable), but in this article, the focus is on the special role of
hyperstrings in it.
The name PISA, by the way, was originally an acronym of Parameter load (i.e.,
a complexity metric), Iteration (i.e., repetition), Symmetry, and Alternation.}
The algorithm takes just symbol strings as input but my point is that
transparallel processing might well be a form of cognitive processing
(van der Helm 2012).
This approach to neural computation has been developed in cognitive science
-- in research on human visual perceptual organization, in particular --
and is communicated here to the artificial intelligence community.
For both domains, the novel observations in this article are (a) that this
classical computing method has the same computing power as that promised by
quantum computers (Feynman 1982), and (b) that it provides a neurally
plausible alternative to the quantum mind hypothesis (Penrose 1989).
Next, to set the stage, five currently relevant ingredients are introduced
briefly.

\subsection{Human visual perceptual organization}

Visual perceptual organization is the neuro-cognitive process that enables us
to perceive scenes as structured wholes consisting of objects arranged in
space (see Fig. 1).
This process may seem to occur effortlessly and we may take it for granted in
daily life, but by all accounts, it must be both complex and flexible.
To give a gist
(following Gray 1999):
For a proximal stimulus, it usually singles out one hypothesis about the
distal stimulus from among a myriad of hypotheses that also would fit the
proximal stimulus (this is the inverse optics problem).
To this end, multiple sets of features at multiple, sometimes overlapping,
locations in a stimulus must be grouped in parallel.
This implies that the process must cope simultaneously with a large number of
possible combinations, which, in addition, seem to interact as if they are
engaged in a stimulus-dependent competition between grouping criteria.
This indicates that the combinatorial capacity of the perceptual organization
process must be high, which, together with its high speed (it completes in
the range of 100--300 ms), reveals its truly impressive nature.

\begin{figure}[t!]

\begin{center}
\includegraphics[width=11cm]{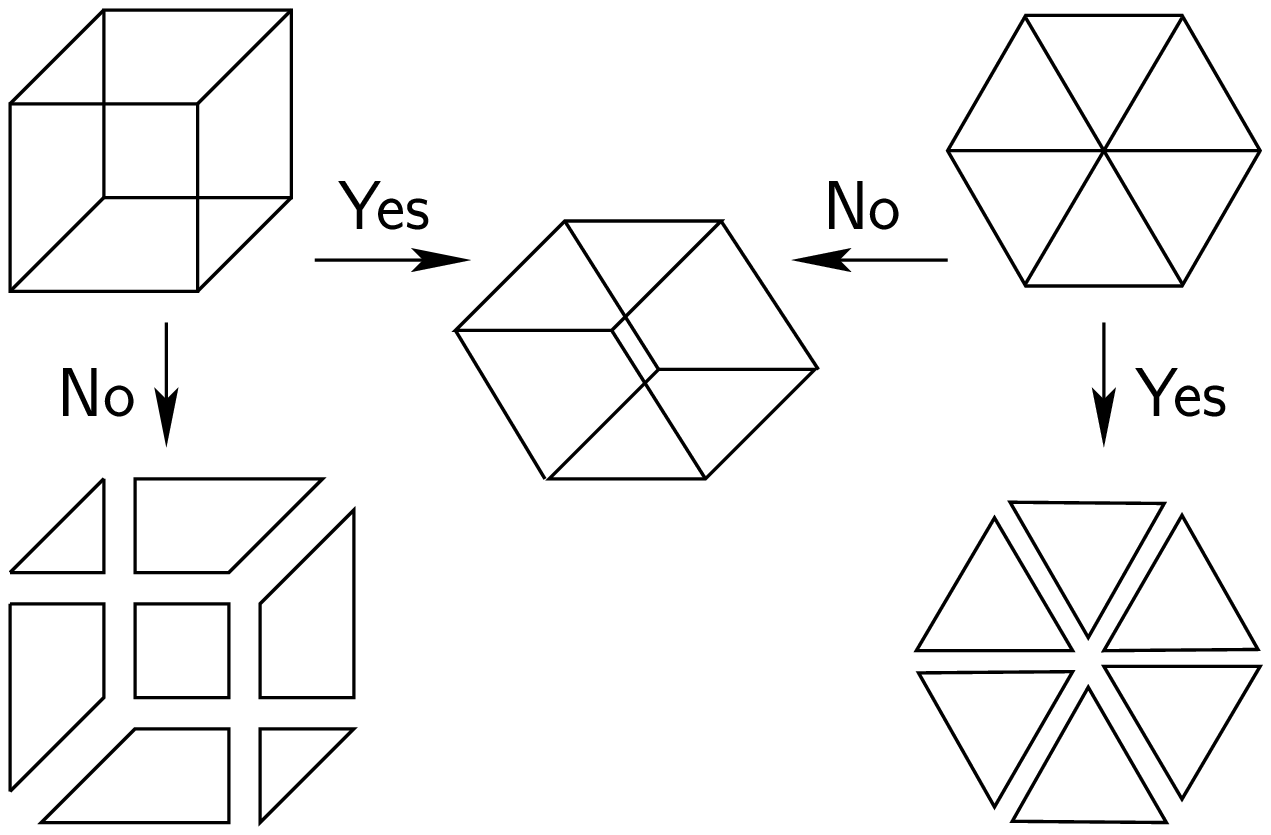}
\end{center}

\small

\textbf{Fig. 1}
Visual perceptual organization.
Both images at the top can be interpreted as 3D cubes and as 2D mosaics, but
as indicated by "Yes" and "No", humans preferably interpret the one at the
left as a 3D cube and the one at the right as a 2D mosaic
(after Hochberg and Brooks 1960)

\normalsize

\vspace{\baselineskip}
\vspace{\baselineskip}

\end{figure}

I think that this cognitive process must involve something additional to
traditionally considered forms of processing
(see also Townsend and Nozawa 1995).
The basic form of processing thought to be performed by the neural network of
the brain is parallel distributed processing (PDP), which typically involves
interacting agents who simultaneously do different things.
However, the brain also exhibits a more sophisticated processing mode, namely,
neuronal synchronization, which involves interacting agents who simultaneously
do the same thing -- think of flash mobs or choirs going from
cacophony to harmony.

\subsection{Neuronal synchronization}

Neuronal synchronization is the phenomenon that neurons, in transient
assemblies, temporarily synchronize their firing activity.
Such assemblies are thought to arise when neurons shift their allegiance to
different groups by altering connection strengths, which may also imply a
shift in the specificity and function of neurons
(Edelman 1987;
Gilbert 1992).
Both theoretically and empirically, neuronal synchronization has been
associated with various cognitive processes and 30--70 Hz gamma-band
synchronization, in particular, has been associated with feature
binding in visual perceptual organization
(Eckhorn et al. 1988;
Gray 1999;
Gray and Singer 1989).

Ideas about the meaning of  gamma-band synchronization are, for instance, that
it binds neurons which, together, represent one perceptual entity
(Milner 1974;
von der Malsburg 1981),
or that it is a marker that an assembly has arrived at a steady state
(Pollen 1999),
or that its strength is an index of the salience of features
(Finkel et al. 1998;
Salinas and Sejnowski 2001),
or that more strongly synchronized assemblies in a visual area in the brain
are locked on more easily by higher areas
(Fries 2005).
These ideas sound plausible, that is, synchronization indeed might reflect a
flexible and efficient mechanism subserving the representation of information,
the regulation of the flow of information, and the storage and retrieval of
information
(Sejnowski and Paulsen 2006;
Tallon-Baudry 2009).

I do not challenge those ideas, but notice that they are about cognitive
factors associated with synchronization rather than about the nature of the
underlying cognitive processes.
In other words, they merely express that synchronization is a manifestation of
cognitive processing -- just as the bubbles in boiling water are a
manifestation of the boiling process
(Bojak and Liley 2007; Shadlen and Movshon 1999).
The question then still is, of course, of what form of cognitive processing it
might be a manifestation.
My stance in this article is that neuronal synchronization might well be a
manifestation of transparallel information processing, which, as I explicate
later on, means that many similar features are processed simultaneously as if
only one feature were concerned.

Apart from the above ideas about the meaning of synchronization, two lines of
research into the physical mechanisms of synchronization are worth mentioning.
First, research using methods from dynamic systems theory (DST) showed that
the occurrence of synchronization and desynchronization in a network depends
crucially on system parameters that regulate interactions between nodes
(see, e.g., Buzs\'aki 2006; Buzs\'aki and Draguhn 2004;
Campbell et al. 1999; van Leeuwen 2007; van Leeuwen et al. 1997).
This DST research is relevant, because it investigates -- at the level of
neurons -- how synchronized assemblies might go in and out of existence.
Insight therein complements research, like that in this article, into what
these assemblies do in terms of cognitive information processing.
Second, the quantum mind hypothesis postulated that quantum mechanical
phenomena, such as quantum entanglement and superposition, are the subneuron
source of neuronal synchronization which, in turn, might underlie
consciousness
(Penrose 1989; Penrose and Hameroff 2011; see also Atmanspacher 2011).
This hypothesis is controversial, mainly because quantum mechanical
phenomena do not seem to last long enough to be useful for neuro-cognitive
processing, let alone for consciousness
(Chalmers 1995; Chalmers 1997; Searle 1997; Seife 2000;
Stenger 1992; Tegmark 2000).
Be that as it may, the quantum mind hypothesis had been inspired by quantum
computing which is a currently relevant form of processing.

\subsection{Quantum computing}

Quantum computing, an idea from physics
(Deutsch and Jozsa 1992;
Feynman 1982),
is often said to be the holy grail of computing: It promises --
rightly or wrongly -- to be exponentially faster than classical computing.
More specifically, the difference between classical computers and quantum
computers is as follows.
Classical computers, on the one hand, work with binary digits ({\it bits})
which each represent either a one or a zero, so that a classical computer with
$N$ bits can be in only one of $2^N$ states at any one time.
Quantum computers, on the other hand, work with quantum bits ({\it qubits})
which each can represent a one, a zero, or any quantum superposition of these
two qubit states.
A quantum computer with $N$ qubits can therefore be in an arbitrary
superposition of $O(2^N)$ states simultaneously.\footnote{\label{big-O}
Formally, for functions $f$ and $g$ defined on the positive integers,
$f$ is $O(g)$ if a constant $C$ and a positive integer $n_0$ exist such that
$f(n) \leq C*g(n)$ for all $n \geq n_0$.
Informally, $f$ then is said to be in the order of magnitude of $g$.}
A final read-out will give one of these states but, crucially, (a) the outcome
of the read-out is affected directly by this superposition, which (b)
effectively means that, until the read-out, all these states can be dealt with
simultaneously as if only one state were concerned.

The latter suggests an extraordinary computing power.
For instance, many computing tasks (e.g., in string searching) require,
for an input the size of $N$, an exhaustive search among $O(2^N)$ candidate
outputs.
A naive computing method, that is, one that processes each of the $O(2^N)$
options separately, may easily require more time than is available in this
universe (van Rooij 2008),
and compared with that, quantum computing promises an $O(2^N)$ reduction in
the amount of work and time needed to complete a task.

However, the idea of quantum computing also needs qualification.
First, it is true that the quest for quantum computers progresses (e.g., by
the finding of Majorana fermions which might serve as qubits;
Mourik et al. 2012),
but there still are obstacles, and thus far, no scalable
quantum computer has been built.
Second, it is true that quantum computing may speed up some computing tasks
(see, e.g., Deutsch and Jozsa 1992;
Grover 1996;
Shor 1994),
but the vast majority of computing tasks cannot benefit from it
(Ozhigov 1999).
This reflects the general tendency that more sophisticated methods have a
more restricted application domain.
Quantum computing, for instance, requires the applicability of unitary
transformations to preserve the coherence of superposed states.
Third, quantum computers are often claimed to be generally superior to
classical computers, that is, faster than or at least as fast as classical
computers for any computing task.
However, there is no proof of that
(Hagar 2011),
and in this article, I in fact challenge this claim.
To put this in a broader perspective, I next review generic forms of
processing.

\subsection{Generic forms of processing}

In computing or otherwise, a traditional distinction is that between serial
and parallel processing.
Serial processing means that subtasks are performed one after the other by one
processor, and parallel processing means that subtasks are performed
simultaneously by different processors.
In addition, however, one may define subserial processing, meaning that
subtasks are performed one after the other by different processors (i.e., one
processor handles one subtask, and another processor handles the next
subtask).
For instance, the whole process at a supermarket checkout is a form of
multi-threading, but more specifically,
(a) the cashiers work in parallel;
(b) each cashier processes customer carts serially; and
(c) the carts are presented subserially by customers.
Compared to serial processing, subserial processing yields no reduction
in the work and time needed to complete an entire task, while parallel
processing yields reduction in time but not in work.
Subserial processing as such is perhaps not that interesting, but taken
together with serial and parallel processing -- as in Fig. 2 -- it calls for
what I dubbed transparallel processing.

Transparallel processing means that subtasks are performed simultaneously by
one processor, that is, as if only one subtask were concerned.
The next pencil selection metaphor may give a gist.
To select the longest pencil from among a number of pencils, one or many
persons could measure the lengths of the pencils in a (sub)serial or parallel
fashion, after which the lengths can be compared to each other.
However, a smarter -- transparallel -- way would be if one person gathers all
pencils in one bundle upright on a table, so that the longest pencil can be
selected in a glance.\footnote{\label{spaghetti}
The pencil selection example is close to the spaghetti metaphor in sorting
(Dewdney 1984) but serves here primarily to illustrate that, in some
cases, items can be gathered in one bin that can be dealt with as if it
comprised only one item (hyperstrings are such bins).}
This example illustrates that, in contrast to (sub)serial and parallel
processing, transparallel processing reduces the work -- and thereby the time
-- needed to complete a task.

\begin{figure}[t!]

\begin{center}
\includegraphics[width=9cm]{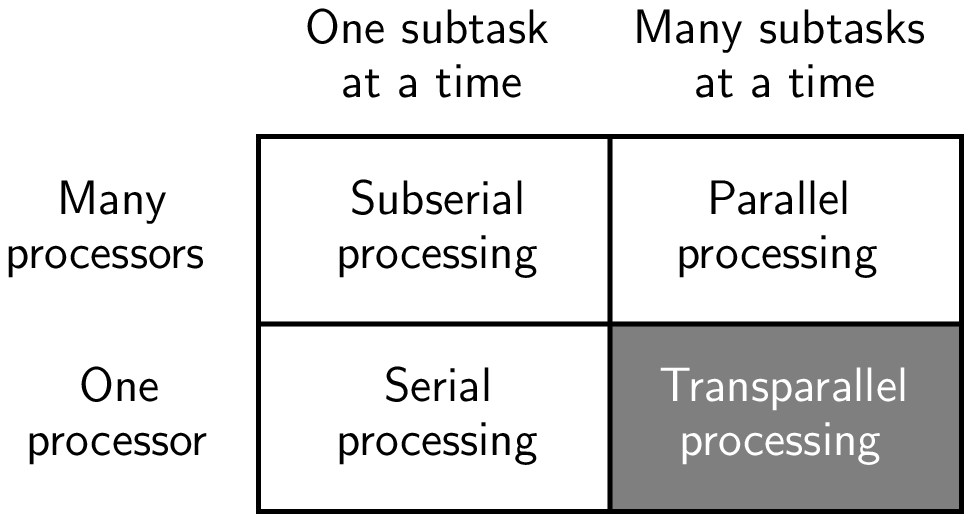}
\end{center}

\small

{\bf Fig. 2}
Generic forms of processing, defined by the number of subtasks performed at a
time (one or many) and the number of processors involved (one or many)

\normalsize

\vspace{\baselineskip}
\vspace{\baselineskip}

\end{figure}

In computing, transparallel processing may seem science-fiction and quantum
computers indeed reflect a prospected hardware method to do
transparallel computing.
However, in Sect.~2, I review an existing software method to do
transparallel computing on single-processor classical computers.
This software method does not fit neatly in existing process taxonomies
(Flynn 1972;
Townsend and Nozawa 1995).
For instance, it escapes the distinction between task parallelism and data
parallelism.
It actually relies on a special kind of distributed representations, which I
next briefly introduce in general.

\subsection{Distributed representations}

In both classical and quantum computing, the term distributed processing is
often taken to refer to a process that is divided over a number of processors.
Also then, it refers more generally (i.e., independently of the
number of processors involved) to a process that operates on a distributed
representation of information.
For instance, in the search for extraterrestrial intelligence (SETI), a
central computer maintains a distributed representation of the sky:
It divides the sky into pieces that are analyzed by different computers which
report back to the central computer.
Furthermore, PDP models in cognitive science often use the brain metaphor
of activation spreading in a network of processors which operate in parallel
to regulate interactions between pieces of information stored at different
places in the network
(e.g., Rumelhart and McClelland 1982).
The usage of distributed representations in SETI is a form of data
parallelism, which, just as task parallelism, reduces the time but not the
work needed to complete an entire task.
In PDP models, however, it often serves to achieve a reduction in work -- and,
thereby, also in time.

Formally, a distributed representation is a data structure that can
be visualized by a graph
(Harary 1994),
that is, by a set of interconnected nodes, in which pieces of information
are represented by the nodes, or by the links, or by both.
One may think of road maps, in which roads are represented by
links between nodes that represent places, so that possibly overlapping
sequences of successive links represent whole routes.
Work-reducing distributed representations come in various flavors, but for $N$
nodes, they typically represent superpositions of $O(2^N)$ wholes by means of
only $O(N^2)$ parts.
To find a specific whole, a process then might confine itself to examining
only the $O(N^2)$ parts.
Well-known examples in computer science are the shortest path method
(Dijkstra 1959)
and methods using suffix trees
(Gusfield 1997)
and deterministic finite automatons
(Hopcroft and Ullman 1979).

In computer science, such classical computing methods are also called smart
methods, because, compared to a naive method that processes each of those
$O(2^N)$ wholes separately, they reduce an exponential $O(2^N)$ amount of work
to a polynomial one -- typically one between $O(N)$ and $O(N^2)$.
Hence, their computing power is between that of naive methods and that of
quantum computing, the latter reducing an exponential $O(2^N)$ amount of work
to a constant $O(1)$ one.
The main points in the remainder of this article now are, first, that
hyperstrings are distributed representations that take classical computing
to the level of quantum computing, and second, that they enable a neurally
plausible alternative to the quantum mind hypothesis.

\section{Classical computing with quantum power}

The idea that the brain represents information in a reduced, distributed,
fashion has been around for a while
(Hinton 1990).
Fairly recently, ideas have emerged -- in both cognitive science and
computer science -- that certain distributed representations allow for
information processing that is mathematically analogous to quantum computing
(Aerts et al. 2009),
or might even allow for classical computing with quantum power
(Rinkus 2012).
Returning in these approaches is that good candidates for that are so-called
holographic reduced representations
(Plate 1991).
To my knowledge, however, these approaches have not yet achieved actual
classical computing with quantum power -- which hyperstrings do achieve.
Formally, hyperstrings differ from holographic reduced representations, but
conceptually, they seem to have something in common: Hyperstrings are
distributed representations of what I called holographic regularities
(van der Helm 1988; van der Helm and Leeuwenberg 1991).
Let me first briefly give some background thereof.

\subsection{Background}

The minimal coding problem, for which hyperstrings enable a solution, arose in
the context of structural information theory (SIT), which is a general theory
of human visual perceptual organization (Leeuwenberg and van der Helm 2013).
In line with the Gestalt law of Pr\"agnanz
(Wertheimer 1912, 1923; K\"ohler 1920; Koffka 1935), it adopts the
simplicity principle, which aims at economical mental representations:
It holds that the simplest organization of a stimulus is the one most likely
perceived by humans (Hochberg and McAlister 1953).
To make quantifiable predictions, SIT developed a formal coding model for
symbol strings, which, in the Appendix, is presented and illustrated in the
form it has since about 1990.
The minimal coding problem thus is the problem to compute guaranteed simplest
codes of strings, that is, codes which -- by exploiting regularities --
specify strings by a minimum number of descriptive parameters.

SIT, of course, does not assume that the human visual system converts visual
stimuli into strings.
Instead, it uses manually obtained strings to represent stimulus
interpretations in the sense that such a string can be read as a series of
instructions to reproduce a stimulus (much like a
computer algorithm is a series of instructions to produce output).
For instance, for a line pattern, a string may represent the sequence of
angles and line segments in the contour.
A stimulus can be represented by various strings, and the string with the
simplest code is taken to reflect the prefered interpretation, with a
hierarchical organization as described by that simplest code.
In other words, SIT assumes that the processing principles, which its formal
model applies to strings, reflect those which the human visual system
applies to visual stimuli.

The regularities considered in SIT's formal coding model are repetition
(juxtaposed repeats), symmetry (mirror symmetry and broken symmetry),
and alternation (nonjuxtaposed repeats).
These are mathematically unique in that they are the only regularities with a
hierarchically transparent holographic nature
(for details, see van der Helm and Leeuwenberg 1991, or its clearer version in
van der Helm 2014).
To give a gist, the string $ababbaba$ exhibits a global symmetry, which can be
coded step by step, each step adding one identity relationship between
substrings -- say, from $aba\ S[(b)]\ aba$ to $ab\ S[(a)(b)]\ ba$, and so on
until $S[(a)(b)(a)(b)]$.
The fact that such a stepwise expansion preserves symmetry illustrates that
symmetry is holographic.
Furthermore, the argument $(a)(b)(a)(b)$ of the resulting symmetry code can be
hierarchically recoded into $2*((a)(b))$, and the fact that this repetition
corresponds unambiguously to the repetition $2*(ab)$ in the original string
illustrates that symmetry is hierarchically transparent.

The formal properties of holography and hierarchical transparency not only
single out repetition, symmetry, and alternation, but are also perceptually
adequate.
They explain much of the human perception of single and combined
regularities in visual stimuli, whether or not perturbed by noise
(for details,
see van der Helm 2014;
van der Helm and Leeuwenberg 1996, 1999, 2004).
For instance, they explain that mirror symmetry and Glass patterns are better
detectable than repetition, and that the detectability of mirror symmetry and
Glass patterns in the presence of noise follows a psychophysical law that
improves on Weber's law
(van der Helm 2010).
Currently relevant is their role in the minimal coding problem -- next,
this is discussed in more detail.

\subsection{Hyperstrings}

At first glance, the minimal coding problem seems to involve just the
detection of regularities in a string, followed by the selection of a
simplest code (see Fig. 3).
However, these are actually the relatively easy parts of the problem -- they
can be solved by traditional computing methods
(van der Helm 2004, 2012, 2014).
Therefore, here, I focus not so much on the entire minimal coding problem, but
rather on its hard part, namely, the problem that the argument of every
detected symmetry or alternation has to be hierarchically recoded before a
simplest code may be selected (repetition does not pose this problem).
As spelled out in the Appendix, this implies that a string of length $N$
gives rise to a superexponential $O(2^{N \log N})$ number of possible
codes.
This has raised doubts about the tractability of minimal coding, and thereby,
about the adequacy of the simplicity principle in perception
(e.g., Hatfield and Epstein 1985).

\begin{figure}[b!]

\vspace{\baselineskip}
\vspace{\baselineskip}

\begin{center}
\includegraphics[width=14cm]{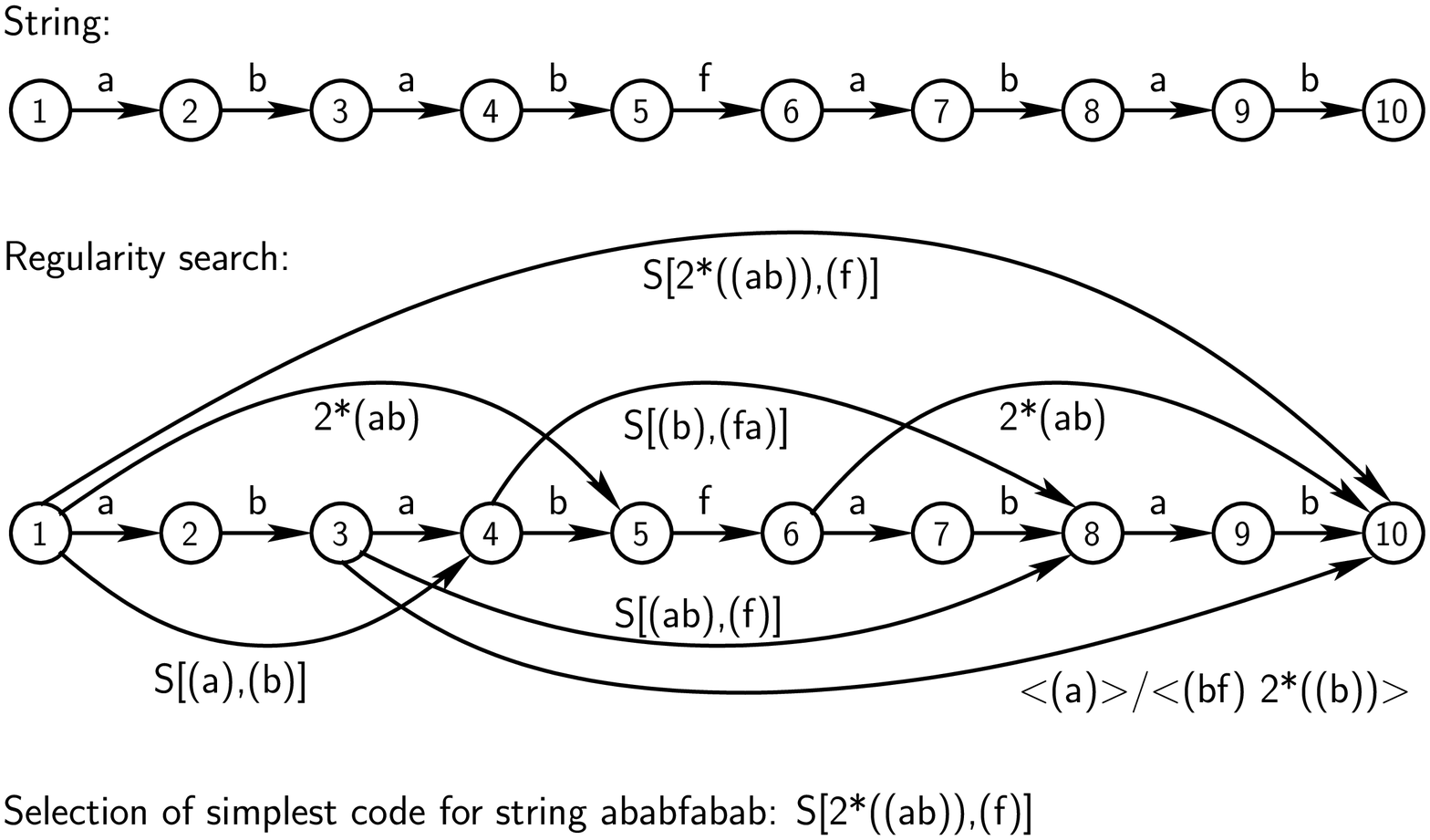}
\end{center}

\small

{\bf Fig. 3}
Minimal coding.
Encoding the string $ababfabab$ involves an exhaustive search for subcodes
capturing regularities in substrings (here, only a few are shown).
These subcodes can be gathered in a directed acyclic graph, in which every
edge represents a substring -- with the complexity of the simplest subcode of
the substring taken as its length, so that the shortest path through the
graph represents the simplest code of the entire string

\normalsize

\end{figure}

Yet, the hard part of minimal coding can be solved, namely, by first
gathering similar regularities in distributed representations
(van der Helm 2004;
van der Helm and Leeuwenberg 1991).
For instance, the string {\it ababfababbabafbaba} exhibits, among others, the
symmetry {\it S[(aba)(b)(f)(aba)(b)]}.
Its argument {\it (aba)(b)(f)(aba)(b)} is represented in Fig. 4a by the path
along vertices 1, 4, 5, 6, 9, and 10.
In fact, Fig. 4a represents, in a distributed fashion, the arguments of
all symmetries into which the string can be encoded.
Because of the holographic nature of symmetry, such a distributed
representation for a string of length $N$ can be constructed in $O(N^2)$
computing time, and represents $O(2^N)$ symmetry arguments.
In Fig. 4b, the same has been done for the string {\it ababfababbabafabab},
but notice the difference: Though the input strings differ only slightly, the
sets $\pi(1,5)$ and $\pi(6,10)$ of substrings represented by the subgraphs on
vertices 1--5 and 6--10 are identical in Fig. 4a but disjoint in Fig. 4b.

\begin{figure}[t!]

\begin{center}
\includegraphics[width=16cm]{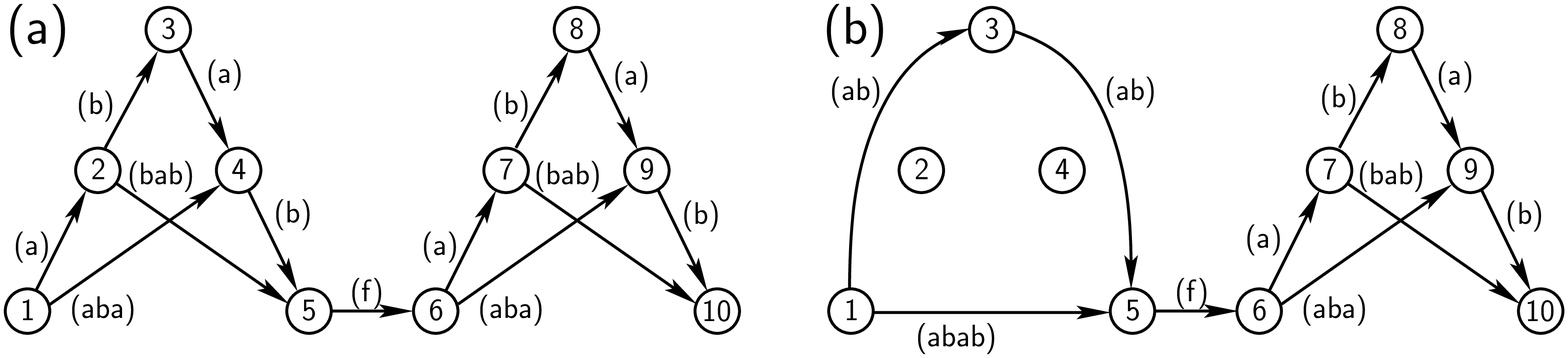}
\end{center}

\small

{\bf Fig. 4}
Distributed representations of similar regularities.
(a) Graph representing the arguments of all symmetries into which the
string {\it ababfababbabafbaba} can be encoded.
(b) Graph representing the arguments of all symmetries into which the
slightly different string {\it ababfababbabafabab} can be encoded.
Notice that sets $\pi(1,5)$ and $\pi(6,10)$ of substrings represented by
the subgraphs on vertices 1--5 and 6--10 are identical in (a) but disjoint
in (b)

\normalsize

\vspace{\baselineskip}
\vspace{\baselineskip}

\end{figure}

The point now is that, if symmetry arguments (or, likewise, alternation
arguments) are gathered this way, then the resulting distributed
representations consist provably of one (as in Figs. 4a and 4b) or more
independent hyperstrings (see van der Helm 2004, 2014, or the Appendix, for
the formal definition of hyperstrings and for the proofs).
As I clarify next, this implies that a single-processor classical computer can
hierarchically recode up to an exponential number of symmetry or alternation
arguments in a transparallel fashion, that is, simultaneously as if only one
argument were concerned.

\begin{figure}[t!]

\begin{center}
\includegraphics[width=12cm]{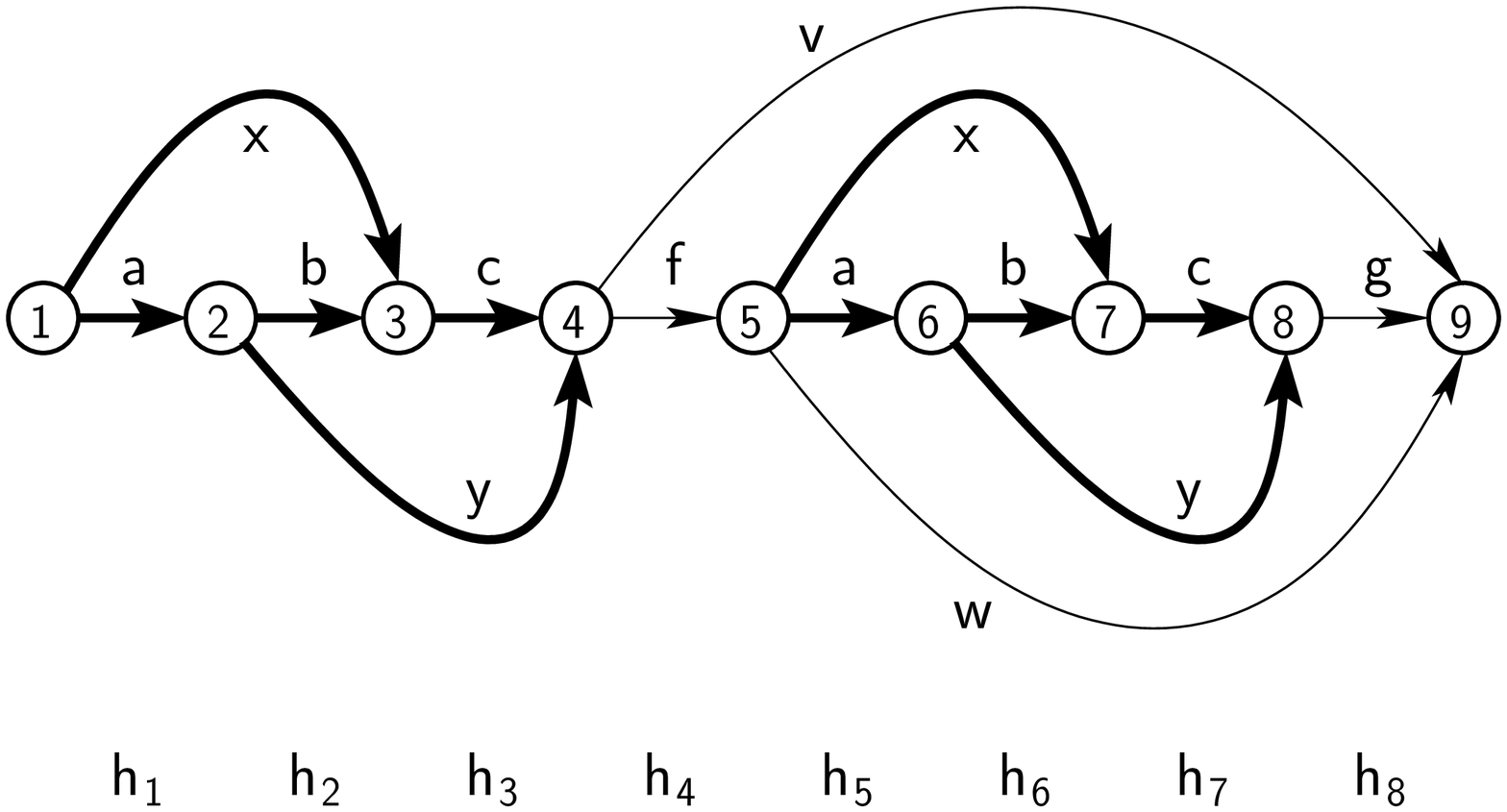}
\end{center}

\small

{\bf Fig. 5}
A hyperstring.
Every path from source (vertex 1) to sink (vertex 9) in the
hyperstring at the top represents a normal string via its edge labels.
The two hypersubstrings indicated by bold edges represent identical substring
sets $\pi(1,4)$ and $\pi(5,8)$, both consisting of the substrings $abc$, $xc$,
and $ay$.
For the string $h_1 ... h_8$ at the bottom, with substrings defined as
corresponding one-to-one to hypersubstrings, this implies
that the substrings $h_1 h_2 h_3$ and $h_5 h_6 h_7$ are identical.
This single identity relationship corresponds in one go to three identity
relationships between substrings in the normal strings, namely, between
the substrings $abc$ in string $abcfabcg$, between the substrings $xc$
in string $xcfxcg$, and between the substrings $ay$ in string $ayfayg$

\normalsize

\vspace{\baselineskip}
\vspace{\baselineskip}

\end{figure}

As illustrated in Fig. 5, a hyperstring is an st-digraph (i.e., a directed
acyclic graph with one source and one sink) with, crucially, one Hamiltonian
path from source to sink (i.e., a path that visits every vertex only once).
Every source-to-sink path in a hyperstring represents some normal
string consisting of some number of elements, but crucially, substring sets
represented by hypersubstrings are either identical or disjoint
(as illustrated in Fig. 4).
This implies that every identity relationship between substrings in one of the
normal strings corresponds to an identity relationship between
hypersubstrings, and that inversely, every identity relationship between
hypersubstrings corresponds in one go to identity relationships between
substrings in several of the normal strings.

The two crucial properties above imply that a hyperstring can be searched for
regularity as if it were a single normal string.
For instance, the hyperstring in Fig. 5 can be treated as if it were a
string $h_1 ... h_8$, with substrings that correspond one-to-one to
hypersubstrings (see Fig. 5).
The latter means that, in this case, the substrings $h_1h_2h_3$ and
$h_5h_6h_7$ are identical, so that the string $h_1 ... h_8$ can be encoded
into, for instance, the alternation
$\langle(h_1h_2h_3)\rangle/\langle (h_4)(h_8)\rangle$.
This alternation thus in fact captures the identity relationship between 
the substring sets $\pi(1,4)$ and $\pi(5,8)$, and thereby it captures in one
go several identity relationships between substrings in different
strings in the hyperstring.
That is, it represents, in one go, alternations in three different strings,
namely:

\begin{quote}
\begin{tabbing}
\qquad $\langle(abc)\rangle/\langle (f)(g)\rangle\quad$ \= in the string \quad $abcfabcg$
\\
\qquad $\langle(xc)\rangle/\langle (f)(g)\rangle\quad$ \> in the string \quad $xcfxcg$
\\
\qquad $\langle(ay)\rangle/\langle (f)(g)\rangle\quad$ \> in the string \quad $ayfayg$
\end{tabbing}
\end{quote}

\begin{figure}[b!]

\vspace{\baselineskip}
\vspace{\baselineskip}

\begin{center}
\includegraphics[width=14cm]{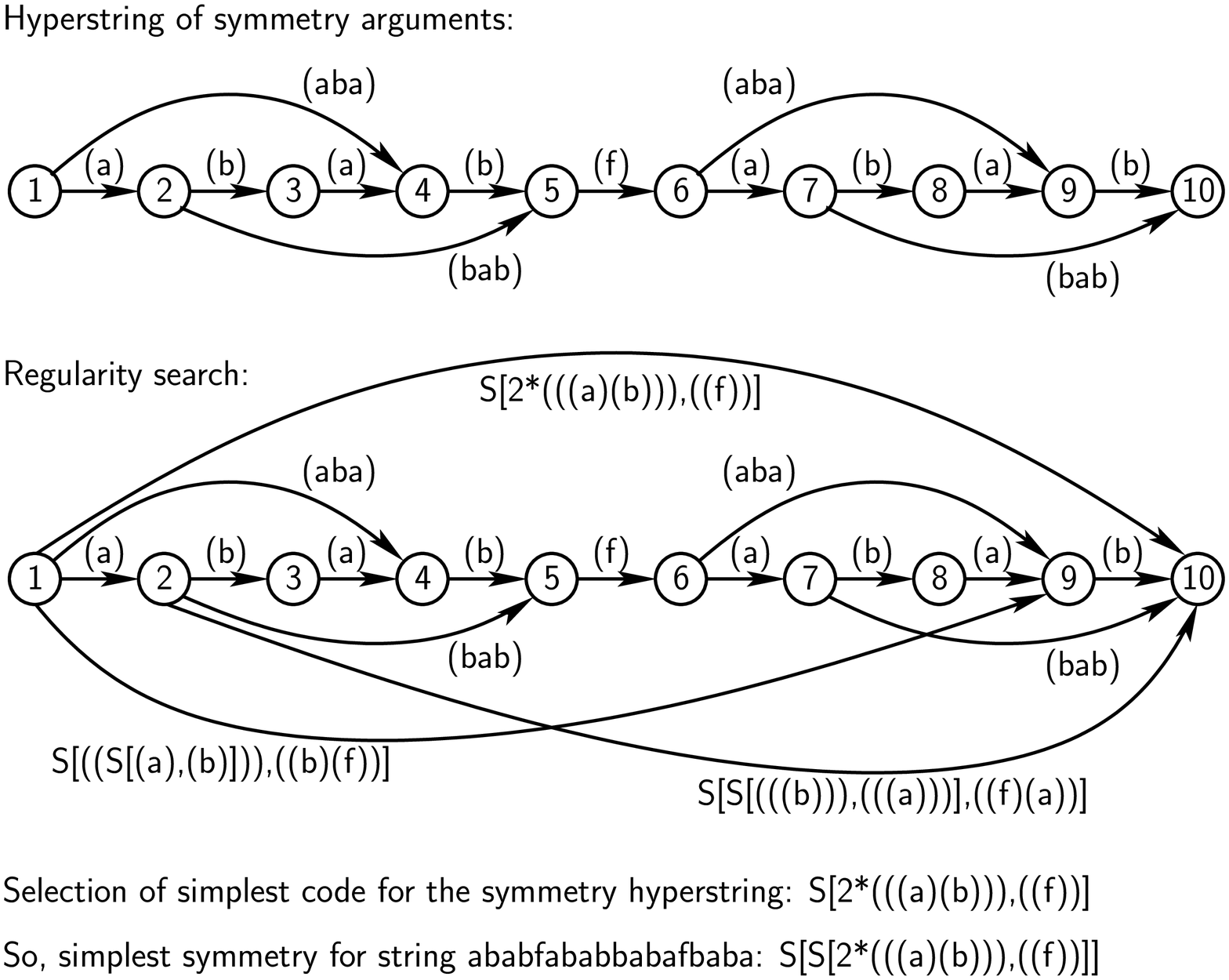}
\end{center}

\small

{\bf Fig. 6}
Hyperstring encoding.
The hyperstring represents the arguments of all symmetries into
which the string {\it ababfababbabafbaba} can be encoded.
The recursive regularity search yields subcodes capturing
regularities in hypersubstrings (here, only a few are shown).

\normalsize

\end{figure}

Returning to hyperstrings of symmetry or alternation arguments, one could say
that they represent up to an exponential number of hypotheses about an input
string, which, as the foregoing illustrates, can be evaluated further
simultaneously.
That is, such a hyperstring can be encoded without having to distinguish
explicitly between the arguments represented in it (see Fig. 6).
At the front end of such an encoding, one, of course, has to establish the
identity relationships between hypersubstrings, but because of the Hamiltonian
path, this can be done in the same way as for a normal string.
At the back end, one, of course, has to select eventually a simplest code, but
by way of the shortest path method (Dijkstra 1959), also this can be done
in the same way as for a normal string.
Thus, in total, recursive hierarchical recoding yields a tree of hyperstrings,
and during the buildup of this tree, parts of it can already be traced back to
select simplest codes of (hyper)substrings -- to select eventually a
simplest code of the entire input string.\footnote{\label{transient}
For a given input string, the tree of hyperstrings and its hyperstrings are
built on the fly, that is, the hyperstrings are transient in that they bind
similar features in the current input only.
This contrasts with standard PDP modeling, which assumes that one fixed
network suffices for many different inputs.}

\subsection{Discussion}

Transparallel processing by hyperstrings has been implemented in the
minimal coding algorithm PISA (see Footnote \ref{pisa}).
PISA, which runs on single-processor classical computers, relies fully on the
mathematical proofs concerning hyperstrings (see the Appendix) and can
therefore be said to provide an algorithmic proof that those mathematical
proofs are correct.
Hence, whereas quantum computers provide a still prospected hardware method to
do transparallel computing, hyperstrings provide an already feasible software
method to do transparallel computing on classical computers.
To be clear, transparallel computing by hyperstrings differs from efficient
classical simulations of quantum computing
(Gottesman 1998).
After all, it implies a truly exponential reduction from $O(2^N)$ subtasks to
one subtask, that is, it implies that $O(2^N)$ symmetry or alternation
arguments can be hierarchically recoded as if only one argument were
concerned.
This implies that it provides classical computers with the same extraordinary
computing power as that promised by quantum computers.
Thus, in fact, it can be said to reflect a novel form of quantum logic
(cf. Dunn et al. 2013),
which challenges the alleged general superiority of quantum computers over
classical computers
(see Hagar 2011).

In general, the applicability of any computing method depends on the
computing task at hand.
This also holds for transparallel computing -- be it by quantum computers or
via hyperstrings by classical computers.
Therefore, I do not dare to speculate on whether, for some tasks, hyperstrings
might give super-quantum power to quantum computers.
Be that as it may, for now, it is true that quantum computers may speed up
some tasks but it is misleading to state in general terms that they will be
exponentially faster than classical computers.
As shown in this section, for at the least one computing task, hyperstrings
provide classical computers with quantum power and it remains to be seen if
quantum computers also can achieve this power for this task.
As I next discuss more speculatively, this application also is relevant to
the question -- in cognitive neuroscience -- of what the computational role
of neuronal synchronization might be.

\section{Human cognitive architecture}

Cognitive architecture, or unified theory of cognition, is a concept from
artificial intelligence research.
It refers to a blueprint for a system that acts like an intelligent system --
taking into account not only its resulting behavior but also physical or more
abstract properties implemented in it
(Anderson 1983; Newell 1990).
Hence, it aims to cover not only competence (what is a system's output?) but
also performance (how does a system arrive at its output?), or in other words,
it aims to unify representations and processes
(Byrne 2012; Langley et al. 2009; Sun 2004; Thagard 2012).
The minimal coding method in the previous section -- developed in the context
of a competence model of human visual perceptual organization --
qualifies as cognitive architecture in the technical sense.
In this section, I investigate if it also complies with ideas about neural
processing in the visual hierarchy in the brain.
From the available neuroscientific evidence, I gather the next picture of
processing in the visual hierarchy.

\subsection{The visual hierarchy}

The neural network in the visual hierarchy is organized with 10--14
distinguishable hierarchical levels (with multiple distinguishable areas
within each level), contains many short-range and long-range connections
(both within and between areas and levels), and can be said to perform a
distributed hierarchical process
(Felleman and van Essen 1991).
This process comprises three neurally intertwined but functionally
distinguishable subprocesses
(Lamme and Roelfsema 2000; Lamme et al. 1998).
As illustrated in the left-hand panel in Fig. 7, these subprocesses are
responsible for
(a) feedforward extraction of, or tuning to, features to which the visual
system is sensitive,
(b) horizontal binding of similar features, and
(c) recurrent selection of different features.
As illustrated in the right-hand panel in Fig. 7, these subprocesses together
yield integrated percepts given by hierarchical organizations (i.e.,
organizations in terms of wholes and parts) of distal stimuli that fit
proximal stimuli.
Attentional processes then may scrutinize these organizations in
a top-down fashion, that is, starting with global structures and, if
required by task and allowed by time, descending to local features
(Ahissar and Hochstein 2004;
Collard and Povel 1982;
Hochstein and Ahissar 2002;
Wolfe 2007).

\begin{figure}[t!]

\begin{center}
\includegraphics[width=16cm]{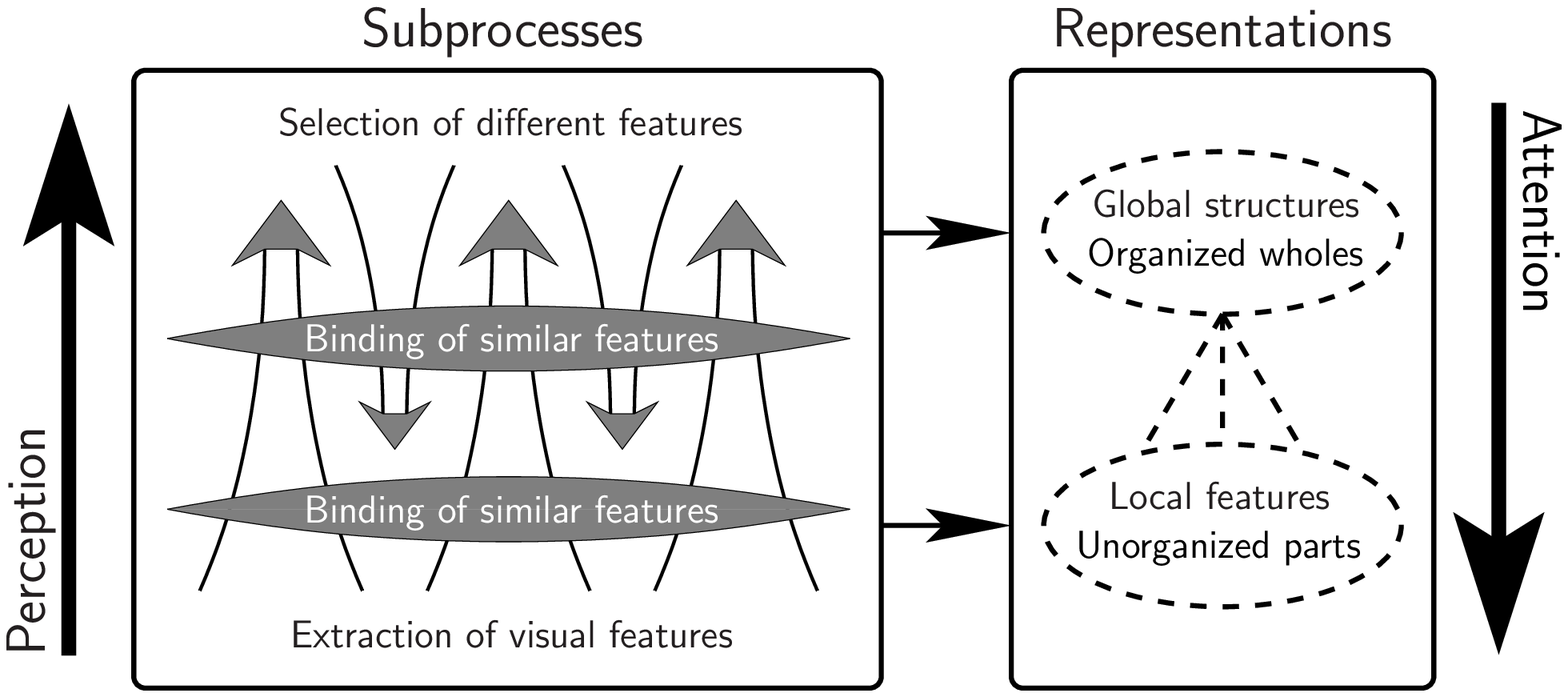}
\end{center}

\small

{\bf Fig. 7}
Processing in the visual hierarchy in the brain.
A stimulus-driven perception process, comprising three neurally intertwined
subprocesses (left-hand panel), yields hierarchical stimulus organizations
(right-hand panel).
A task-driven attention process then may scrutinize
these hierarchical organizations in a top-down fashion

\normalsize

\vspace{\baselineskip}
\vspace{\baselineskip}

\end{figure}

Hence, whereas perception logically processes parts before wholes, the
top-down attentional scrutiny of hierarchical organizations implies that
wholes are experienced before parts.
Thus, this combined action of perception and attention explains the dominance
of wholes over parts, as postulated in early twentieth century Gestalt
psychology (Wertheimer 1912, 1923; K\"ohler 1920; Koffka 1935)
and as confirmed later in a range of behavioral studies (for a review,
see Wagemans et al. 2012).
This dominance also has been specified further by notions such as
global precedence
(Navon 1977),
configural superiority
(Pomerantz et al. 1977),
primacy of holistic properties
(Kimchi 2003),
and superstructure dominance
(Leeuwenberg and van der Helm 1991, Leeuwenberg et al. 1994).

Furthermore, notice that the combination of feedforward extraction and
recurrent selection in perception is like a fountain under increasing
water pressure:
As the feedforward extraction progresses along ascending connections, each
passed level in the visual hierarchy forms the starting point of integrative
recurrent processing along descending connections
(see also VanRullen and Thorpe 2002).
This yields a gradual buildup from partial percepts at lower levels in the
visual hierarchy to complete percepts near its top end.
This gradual buildup takes time, so, it leaves room for attention to intrude,
that is, to modulate things before a percept has completed
(Lamme and Roelfsema 2000;
Lamme et al. 1998).
However, I think that, by then, the perceptual organization process already
has done much of its integrative work (Gray 1999; Pylyshyn 1999),
because, as I sustain next, its speed is high due to, in particular,
transparallel feature processing.

\subsection{The transparallel mind hypothesis}

The perceptual subprocess of feedforward extraction is reminiscent of the
neuroscientific idea that, going up in the visual hierarchy, neural cells
mediate detection of increasingly complex features
(Hubel and Wiesel 1968).
Furthermore, the subprocess of recurrent selection is reminiscent of the
connectionist idea that a standard PDP process of activation spreading in the
brain's neural network yields percepts represented by stable patterns of
activation
(Churchland 1986).
While I think that these ideas capture relevant aspects, I also think that
they are not yet sufficient to account for the high combinatorial capacity and
speed of the human perceptual organization process.
I think that, to this end, the subprocess of horizontal binding is crucial.
This subprocess may be relatively underexposed in neuroscience, but may well
be the neuronal counterpart of the regularity extraction operations, which, in
representational theories like SIT, are proposed to obtain structured mental
representations.

In this respect, notice that the minimal coding method in Sect.~2
relies on three intertwined subprocesses, which correspond to the three
neurally intertwined subprocesses in the visual hierarchy (see Fig. 8).
In the minimal coding method for strings, the formal counterpart of
feedforward extraction is a search for identity relationships between
substrings, by way of an $O(N^2)$ all-substrings identification method akin to
using suffix trees (van der Helm 2014).
Furthermore, it implements recurrent selection by way of the $O(N^3)$
all-pairs shortest path method (Cormen et al. 1994),
which is a distributed processing method that simulates PDP
(van der Helm 2004, 2012, 2014).
Currently most relevant, it implements horizontal binding by gathering similar
regularities in hyperstrings, which allows them to be recoded hierarchically
in a transparallel fashion.

\begin{figure}[t!]

\begin{center}
\includegraphics[width=16cm]{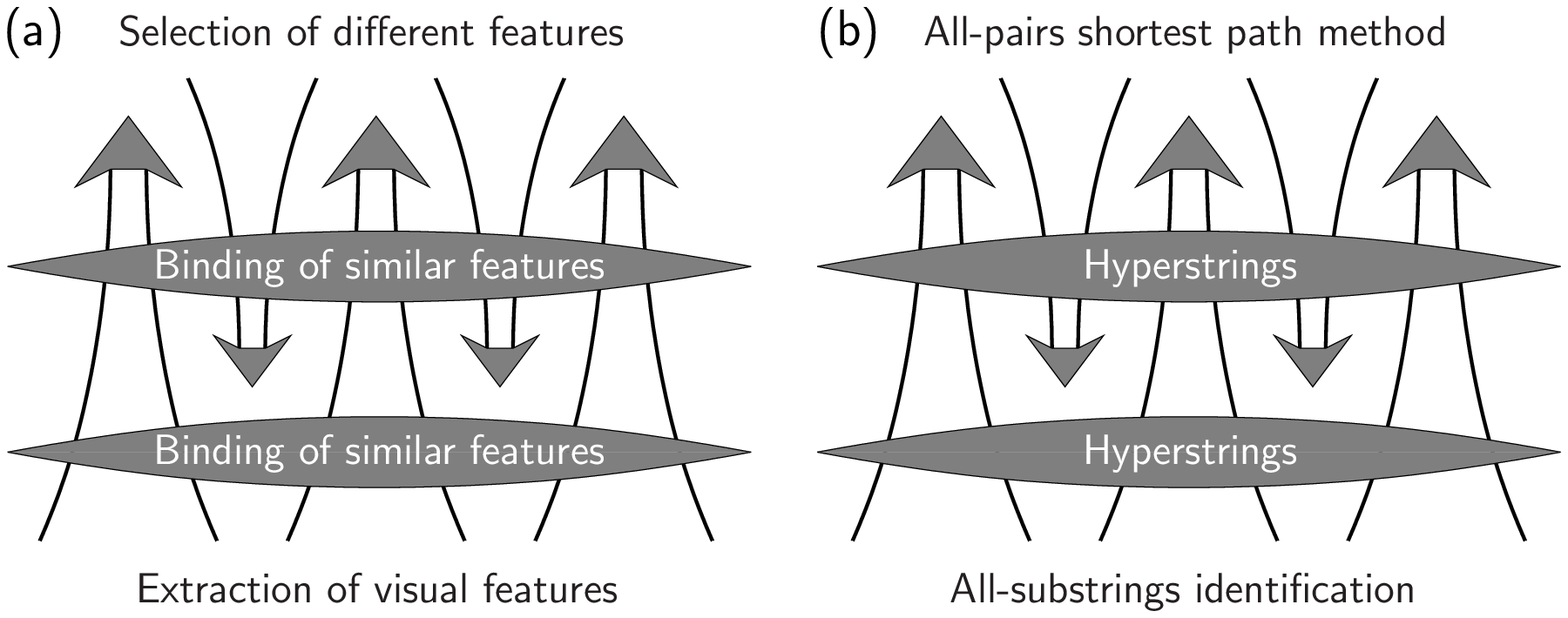}
\end{center}

\small

\noindent
\textbf{Fig. 8}
(a) The three intertwined perceptual subprocesses in the visual hierarchy in
the brain.
(b) The three corresponding subprocesses implemented in the minimal coding
method

\normalsize

\vspace{\baselineskip}
\vspace{\baselineskip}

\end{figure}

This correspondence between three neurally intertwined subprocesses in the
visual hierarchy in brain and three algorithmically intertwined subprocesses
in the minimal coding method is, to me, more than just a nice parallelism.
Epistemologically, it substantiates that knowledge about neural mechanisms and
knowledge about cognitive processes can be combined fruitfully to gain deeper
insights into the workings of the brain (Marr 1982/2010).
Furthermore, ontologically relevant, one of its resulting deeper insights is
that transparallel processing might well be an actual form of processing in the
brain.
That is, horizontal binding of similar features is, in the visual hierarchy in
the brain, thought to be mediated by transient neural assemblies, which signal
their presence by synchronization of the neurons involved (Gilbert 1992).
Synchronization is more than standard PDP and the hyperstring implementation
of horizontal binding in the minimal coding method -- which enables
transparallel processing -- therefore leads me to the following hypothesis
about the meaning of neuronal synchronization.

\begin{quote}
{\it The transparallel mind hypothesis}
\\
Neuronal synchronization is a manifestation of cognitive processing of
similar features in a transparallel fashion.
\end{quote}

\noindent
This hypothesis is, of course, speculative -- but notice that it is based on 
(a) a perceptually adequate and neurally plausible model of the combined
action of human perception and attention, and
(b) a feasible form of classical computing with quantum power.
Next, this hypothesis is discussed briefly in a broader, historical, context.

\subsection{Discussion}

In 1950, theoretical physicist Richard Feynman and cognitive psychologist
Julian Hochberg met in the context of a colloquium they gave.
Feynman would become famous for his work on quantum electrodynamics, and
Hochberg was developing the idea that, among all possible organizations of a
visual stimulus, the simplest one is most likely to be the one perceived by
humans.
They discussed parallels between their ideas and concluded that quantum-like
cognitive processing might underlie such a simplicity principle in human
visual perceptual organization (Hochberg 2012).
This fits in with the long-standing idea that cognition is a dynamic
self-organizing process
(Attneave 1982; Kelso 1995; Koffka 1935; Lehar 2003).
For instance, Hebb (1949) put forward the idea of phase sequence, that is, the
idea that thinking is the sequential activation of sets of neural assemblies.
Furthermore, Rosenblatt (1958) and Fukushima (1975) proposed small artifical
networks -- called perceptrons and cognitrons, respectively -- as formal
counterparts of cognitive processing units.
More recently, the idea arose that actual cognitive processing units
-- or, as I call them, gnosons (i.e., fundamental particles of cognition) --
are given by transient neural assemblies defined by synchronization of the
neurons involved
(Buzs\'aki 2006;
Fingelkurts and Fingelkurts 2001, 2004;
Finkel et al. 1998).

In this article, I expanded on such thinking, by taking hyperstrings as formal
counterparts of gnosons, and by proposing that neuronal synchronization is a
manifestion of transparallel information processing.
Because transparallel processing by hyperstrings provides classical computers
with quantum power, it strengthens ideas that quantum-like cognitive
processing does not have to rely on actual quantum mechanical phenomena at the
subneuron level
(de Barros and Suppes 2009;
Suppes et al. 2012;
Townsend and Nozawa 1995;
Vassilieva et al. 2011).
As I argued, it might rely on interactions at the level of neurons.
It remains to be seen if the transparallel mind hypothesis is tenable too for
synchronization outside the "visual" gamma band, but for one thing, it
accounts for the high combinatorial capacity and speed of human visual
perceptual organization.

\section{Conclusion}

Complementary to DST research on how synchronized neural
assemblies might go in and out of existence due to interactions between
neurons, the transparallel mind hypothesis expresses the representational idea
that neuronal synchronization mediates transparallel information processing.
This idea was inspired by a classical computing method with quantum
power, namely transparallel processing by hyperstrings, which allows up to an
exponential number of similar features to be processed simultaneously as if
only one feature were concerned.
This neurocomputational account thus strengthens ideas that mind is
mediated by transient neural assemblies constituting flexible cognitive
architecture between the relatively rigid level of neurons and the still
elusive level of consciousness.

\section*{Acknowledgments}

I am grateful to Julian Hochberg and Jaap van den Herik for valuable comments
on earlier drafts.
This research was supported by Methusalem grant METH/08/02 awarded to
Johan Wagemans (www.gestaltrevision.be).

\newpage

\section*{References}

\begin{list}{}{\setlength{\leftmargin}{4mm}
               \setlength{\itemsep}{-0.5mm}
               \setlength{\itemindent}{-4mm}}

\item
Aerts D, Czachor M, De Moor B (2009)
Geometric analogue of holographic reduced representation.
J Math Psychol 53:389--398.
doi: 10.1016/j.jmp.2009.02.005

\item
Ahissar M, Hochstein S (2004)
The reverse hierarchy theory of visual perceptual learning.
Trends Cogn Sci 8:457--464.
doi: 10.1016/j.tics.2004.08.011

\item
Anderson JR (1983)
The architecture of cognition.
Harvard University Press, Cambridge

\item
Atmanspacher H (2011)
Quantum approaches to consciousness.
In: Zalta EN (ed)
The Stanford Encyclopedia of Philosophy.
\\
http://plato.stanford.edu/archives/sum2011/entries/qt-consciousness.
\\
Accessed 10 January 2015

\item
Attneave F (1982)
Pr\"agnanz and soap-bubble systems: A theoretical exploration.
In: Beck J (ed)
Organization and representation in perception.
Erlbaum, Hillsdale, pp 11--29

\item
Bojak I, Liley DTJ (2007)
Self-organized 40 Hz synchronization in a physiological theory of EEG.
Neurocomputing 70:2085--2090.
doi: 10.1016/j.neucom.2006.10.087

\item
Buzs\'aki G (2006)
Rhythms of the brain.
Oxford University Press, New York

\item
Buzs\'aki G, Draguhn A (2004)
Neuronal oscillations in cortical networks.
Science 304:1926--1929.
doi: 10.1126/science.1099745

\item
Byrne MD (2012)
Unified theories of cognition.
WIREs Cogn Sci 3:431--438.
\\
doi: 10.1002/wcs.1180

\item
Campbell SR, Wang DL, Jayaprakash C (1999)
Synchrony and desynchrony in integrate-and-fire oscillators.
Neural Comput 11:1595--1619.
doi: 10.1109/IJCNN.1998.685998

\item
Chalmers DJ (1995)
Facing up to the problem of consciousness.
J Conscious Stud 2:200--219.
doi: 10.1093/acprof:oso/9780195311105.003.0001

\item
Chalmers DJ (1997)
The conscious mind: In search of a fundamental theory.
Oxford University Press, Oxford

\item
Churchland PS (1986)
Neurophilosophy.
MIT Press, Cambridge

\item
Collard RF, Povel DJ (1982)
Theory of serial pattern production: Tree traversals.
Psychol Rev 89:693--707.
doi: 10.1037/0033-295X.89.6.693

\item
Cormen TH, Leiserson CE, Rivest RL (1994)
Introduction to algorithms.
MIT Press, Cambridge

\item
de Barros JA, Suppes P (2009)
Quantum mechanics, interference, and the brain.
J Math Psychol 53:306--313.
doi: 10.1016/j.jmp.2009.03.005

\item
Deutsch D, Jozsa R (1992)
Rapid solutions of problems by quantum computation.
Proc R Soc Lond A 439:553--558.
doi: 10.1098/rspa.1992.0167

\item
Dewdney AK (1984)
On the spaghetti computer and other analog gadgets for problem solving.
Sci Am 250:19--26.

\item
Dijkstra EW (1959)
A note on two problems in connexion with graphs.
Numer Math 1:269--271.
doi: 10.1007/BF01386390

\item
Dunn JM, Moss LS, Wang Z (2013)
The third life of quantum logic: Quantum logic inspired by quantum computing.
J Philos Log 42:443--459.
doi: 10.1007/s10992-013-9273-7

\item
Eckhorn R, Bauer R, Jordan W, Brosch M, Kruse W, Munk M, Reitboeck HJ (1988)
Coherent oscillations: A mechanisms of feature linking in the visual cortex?
Biol Cybern 60:121--130.
doi: 10.1007/BF00202899

\item
Edelman GM (1987)
Neural darwinism: The theory of neuronal group selection.
Basic Books, New York

\item
Felleman DJ, van Essen DC (1991)
Distributed hierarchical processing in the primate cerebral cortex.
Cereb Cortex 1:1--47.
doi: 10.1093/cercor/1.1.1

\item
Feynman R (1982)
Simulating physics with computers.
Intern J Theor Phys 21:467--488.
doi: 10.1007/BF02650179

\item
Fingelkurts An.A, Fingelkurts, Al.A (2001)
Operational architectonics of the human brain biopotential field: Towards
solving the mind-brain problem.
Brain and Mind 2:261--296.
doi: 10.1023/A:1014427822738

\item
Fingelkurts An.A, Fingelkurts Al.A (2004)
Making complexity simpler: Multivariability and metastability in the brain.
Intern J Neurosci 114:843--862.
\\
doi: 10.1080/00207450490450046

\item
Finkel LH, Yen S-C, Menschik ED (1998)
Synchronization: The computational currency of cognition.
In: Niklasson L, Boden M, Ziemke T (eds)
ICANN 98 Proc 8th Intern Conf Artif Neural Netw.
Springer-Verlag, New York

\item
Flynn MJ (1972)
Some computer organizations and their effectiveness.
IEEE Transact Comput C-21:948--960.
doi: 10.1109/TC.1972.5009071

\item
Fries P (2005)
A mechanism for cognitive dynamics: Neuronal communication through neuronal
coherence.
Trends Cogn Sci 9:474--480.
doi: 10.1016/j.tics.2005.08.011

\item
Fukushima K (1975)
Cognitron: A self-organizing multilayered neural network.
Biol Cybern 20:121--136.
doi: 10.1007/BF00342633

\item
Gilbert CD (1992)
Horizontal integration and cortical dynamics.
Neuron 9:1--13.
doi: 10.1016/0896-6273(92)90215-Y

\item
Gottesman D (1999)
The Heisenberg representation of quantum computers.
In: Corney SP, Delbourgo R, Jarvis PD (eds)
Group22: Proc XXII Intern Colloq Group Theor Methods Phys.
International Press, Cambridge, pp 32-43.
arXiv: quant-ph/9807006

\item
Gray CM (1999)
The temporal correlation hypothesis of visual feature integration:
Still alive and well.
Neuron 24:31--47.
doi: 10.1016/S0896-6273(00)80820-X

\item
Gray CM, Singer W (1989)
Stimulus-specific neuronal oscillations in orientation columns of cat visual
cortex.
Proc Natl Acad Sci USA 86:1698--1702.
doi: 10.1073/pnas.86.5.1698

\item
Grover LK (1996)
A fast quantum mechanical algorithm for database search.
Proc 28th Ann ACM Symp Theor Comput, pp 212--219.
arXiv: quant-ph/9605043

\item
Gusfield D (1997)
Algorithms on strings, trees, and sequences.
Cambridge University Press, Cambridge

\item
Hagar A (2011)
Quantum computing.
In: Zalta EN (ed)
The Stanford Encyclopedia of Philosophy.
http://plato.stanford.edu/archives/spr2011/entries/qt-quantcomp.
Accessed 10 January 2015

\item
Harary F (1994)
Graph theory.
Addison-Wessley, Reading

\item
Hatfield GC, Epstein W (1985)
The status of the minimum principle in the theoretical analysis of
visual perception.
Psychol Bull 97:155--186.
doi: 10.1037/0033-2909.97.2.155

\item
Hebb DO (1949)
The organization of behavior.
Wiley and Sons, New York

\item
Hinton GE (1990)
Mapping part-whole hierarchies into connectionist networks.
Artif Intel 46:47--75.
doi: 10.1016/0004-3702(90)90004-J

\item
Hochberg JE (June 4, 2012).
Personal communication

\item
Hochberg JE, Brooks V (1960)
The psychophysics of form: Reversible-perspective drawings of spatial objects.
Am J Psychol 73:337--354.
doi: 10.2307/1420172

\item
Hochberg JE, McAlister E (1953)
A quantitative approach to figural "goodness".
J Exp Psychol 46:361--364.
doi: 10.1037/h0055809

\item
Hochstein S, Ahissar M (2002)
View from the top: Hierarchies and reverse hierarchies in the visual system.
Neuron 36:791--804.
doi: 10.1016/S0896-6273(02)01091-7

\item
Hopcroft JE, Ullman JD (1979)
Introduction to automata theory, languages, and computation.
Addison-Wesley, Reading

\item
Hubel DH, Wiesel TN (1968)
Receptive fields and functional architecture of monkey striate cortex.
J Physiol Lond 195:215--243

\item
Kelso JAS (1995)
Dynamic patterns: the self-organization of brain and behavior.
MIT Press, Cambridge

\item
Kimchi R (2003)
Relative dominance of holistic and component properties in the perceptual
organization of visual objects.
In: Peterson MA, Rhodes G (eds)
Perception of faces, objects, and scenes: Analytic and holistic processes.
Oxford University Press, New York, pp 235--263

\item
Koffka K (1935)
Principles of Gestalt psychology.
Routledge and Kegan Paul, London

\item
K\"ohler, W. (1920).
{\it Die physischen Gestalten in Ruhe und im station\"aren Zustand}
[Static and stationary physical shapes].
Braunschweig, Germany: Vieweg

\item
Lamme VAF, Roelfsema PR (2000)
The distinct modes of vision offered by feedforward and recurrent processing.
Trends Neurosci 23:571--579.
doi: 10.1016/S0166-2236(00)01657-X

\item
Lamme VAF, Sup\`er H, Spekreijse H (1998)
Feedforward, horizontal, and feedback processing in the visual cortex.
Curr Opin Neurobiol 8:529--535.
doi: 10.1016/S0959-4388(98)80042-1

\item
Langley P, Laird JE, Rogers S (2009)
Cognitive architectures: Research issues and challenges.
Cogn Syst Res 10:141--160.
doi: 10.1016/j.cogsys.2006.07.004

\item
Leeuwenberg ELJ, van der Helm PA (1991)
Unity and variety in visual form.
Perception 20:595--622.
doi: 10.1068/p200595

\item
Leeuwenberg ELJ, van der Helm PA (2013)
Structural information theory: The simplicity of visual form.
Cambridge University Press, Cambridge

\item
Leeuwenberg ELJ, van der Helm PA, van Lier RJ (1994)
From geons to structure: A note on object classification.
Perception 23:505--515.
doi: 10.1068/p230505

\item
Lehar S (2003)
Gestalt isomorphism and the primacy of the subjective conscious experience: A
Gestalt bubble model.
Behav Brain Sci 26:375--444.
\\
doi: 10.1017/S0140525X03000098

\item
Marr D (2010)
Vision.
MIT Press, Cambridge
(Original work published 1982 by Freeman)

\item
Milner P (1974)
A model for visual shape recognition.
Psychol Rev 81:521--535.
doi: 10.1037/h0037149

\item
Mourik V, Zuo K, Frolov SM, Plissard SR, Bakkers EPAM, Kouwenhoven LP (2012)
Signatures of majorana fermions in hybrid superconductor-semiconductor
nanowire devices.
Science 336:1003--1007.
doi: 10.1126/science.1222360

\item
Navon D (1977)
Forest before trees: The precedence of global features in visual perception.
Cogn Psychol 9:353--383.
doi: 10.1016/0010-0285(77)90012-3

\item
Newell A (1990)
Unified theories of cognition.
Harvard University Press, Cambridge

\item
Ozhigov Y (1999)
Quantum computers speed up classical with probability zero.
Chaos Solitons Fractals 10:1707--1714.
doi: 10.1016/S0960-0779(98)00226-4

\item
Penrose R (1989)
The emperor's new mind: Concerning computers, minds and the laws of physics.
Oxford University Press, Oxford

\item
Penrose R, Hameroff S (2011)
Consciousness in the universe: Neuroscience, quantum space-time geometry and
Orch OR theory.
J Cosmol 14.
\\
http://journalofcosmology.com/Consciousness160.html.
Accessed 10 January 2015

\item
Plate TA (1991)
Holographic reduced representations: Convolution algebra for compositional
distributed representations.
In: Mylopoulos J, Reiter R (eds)
Proc 12th Intern Joint Conf Artif Intel.
Morgan Kaufmann, San Mateo, pp 30--35

\item
Pollen DA (1999)
On the neural correlates of visual perception.
Cereb Cortex 9:4--19.
doi: 10.1093/cercor/9.1.4

\item
Pomerantz JR, Sager LC, Stoever RJ (1977)
Perception of wholes and their component parts: Some configural superiority
effects.
J Exp Psychol: Human Percept Perform 3:422--435.
doi: 10.1037/0096-1523.3.3.422

\item
Pylyshyn ZW (1999)
Is vision continuous with cognition?
The case of impenetrability of visual perception.
Behav Brain Sci 22:341--423

\item
Rinkus GJ (2012)
Quantum computing via sparse distributed representations.
NeuroQuantology 10:311--315.
doi: 10.14704/nq.2012.10.2.507

\item
Rosenblatt F (1958)
The perceptron: A probabilistic model for information storage and organization
in the brain.
Psychol Rev, 65:386--408.
doi: 10.1037/h0042519

\item
Rumelhart DE, McClelland JL (1982)
An interactive activation model of context effects in letter perception:
Part 2. The contextual enhancement effect and some tests and extensions of
the model.
Psychol Rev 89:60--94.
doi: 10.1037/0033-295X.89.1.60

\item
Salinas E, Sejnowski TJ (2001)
Correlated neuronal activity and the flow of neural information.
Nature Rev Neurosci 2:539--550.
doi: 10.1038/35086012

\item
Searle JR (1997)
The mystery of consciousness.
The New York Review of Books, New York

\item
Seife C (2000)
Cold numbers unmake the quantum mind.
Science 287:791.
\\
doi: 10.1126/science.287.5454.791

\item
Sejnowski TJ, Paulsen O (2006)
Network oscillations: Emerging computational principles.
J Neurosci 26:1673--1676.
doi: 10.1523/JNEUROSCI.3737-05d.2006

\item
Shadlen MN, Movshon JA (1999)
Synchrony unbound: A critical evaluation of the temporal binding hypothesis.
Neuron 24:67--77.
doi: 10.1016/S0896-6273(00)80822-3

\item
Shor PW (1994)
Algorithms for quantum computation: Discrete logarithms and factoring.
In: Goldwasser S (ed),
Proc 35nd Ann Symp Found Comput Sci.
IEEE Computer Society Press, Washington, pp 124--134

\item
Stenger V (1992)
The myth of quantum consciousness.
The Humanist 53:13--15

\item
Sun R (2004)
Desiderata for cognitive architectures.
Philos Psychol 3:341--373.
doi: 10.1080/0951508042000286721

\item
Suppes P, de Barros JA, Oas G (2012)
Phase-oscillator computations as neural models of stimulus-response
conditioning and response selection.
J Math Psychol 56:95--117.
doi: 10.1016/j.jmp.2012.01.001

\item
Tallon-Baudry C (2009)
The roles of gamma-band oscillatory synchrony in human visual cognition.
Front Biosci 14:321--332.
doi: 10.2741/3246

\item
Tegmark M (2000)
Importance of quantum decoherence in brain processes.
Phys Rev E 61:4194--4206.
doi: 10.1103/PhysRevE.61.4194

\item
Thagard P (2012)
Cognitive architectures.
In: Frankish K, Ramsay W (eds)
The Cambridge handbook of cognitive science,
Cambridge University Press, Cambridge, pp 50--70

\item
Townsend JT, Nozawa G (1995)
Spatio-temporal properties of elementary perception: An investigation of
parallel, serial, and coactive theories.
J Math Psychol 39:321--359.
doi: 10.1006/jmps.1995.1033

\item
van der Helm PA (1988)
Accessibility and simplicity of visual structures.
PhD Dissertation, Radboud University Nijmegen

\item
van der Helm PA (2004)
Transparallel processing by hyperstrings.
Proc Natl Acad Sci USA 101:10862--10867.
doi: 10.107:pnas.0403402101

\item
van der Helm PA (2010)
Weber-Fechner behaviour in symmetry perception?
Atten Percept Psychophys 72:1854--1864.
doi: 10.3758/APP.72.7.1854

\item
van der Helm PA (2012)
Cognitive architecture of perceptual organization: From neurons to gnosons.
Cogn Proc 13:13--40.
doi: 10.1007/s10339-011-0425-9

\item
van der Helm PA (2014)
Simplicity in vision:
A multidisciplinary account of perceptual organization.
Cambridge University Press, Cambridge

\item
van der Helm PA, Leeuwenberg ELJ (1991)
Accessibility, a criterion for regularity and hierarchy in visual pattern
codes.
J Math Psychol 35:151--213.
doi: 10.1016/0022-2496(91)90025-O

\item
van der Helm PA, Leeuwenberg ELJ (1996)
Goodness of visual regularities: A nontransformational approach.
Psychol Rev 103:429--456.
doi: 10.1037/0033-295X.103.3.429

\item
van der Helm PA, Leeuwenberg ELJ (1999)
A better approach to goodness: Reply to Wagemans (1999).
Psychol Rev 106:622--630.
doi: 10.1037/0033-295X.106.3.622

\item
van der Helm PA, Leeuwenberg ELJ (2004)
Holographic goodness is not that bad: Reply to Olivers, Chater, and
Watson (2004).
Psychol Rev 111:261--273.
doi: 10.1037/0033-295X.111.1.261

\item
van Leeuwen C (2007)
What needs to emerge to make you conscious?
J Conscious Stud 14:115--136

\item
van Leeuwen C, Steyvers M, Nooter M (1997)
Stability and intermittency in large-scale coupled oscillator models for
perceptual segmentation.
J Math Psychol 41:319--344.
doi: 10.1006/jmps.1997.1177

\item
van Rooij I (2008)
The tractable cognition thesis.
Cogn Sci 32:939--984.
\\
doi: 10.1080/03640210801897856

\item
VanRullen R, Thorpe SJ (2002)
Surfing a spike wave down the ventral stream.
Vis Res 42:2593--2615.
doi: 10.1016/S0042-6989(02)00298-5

\item
Vassilieva E, Pinto G, de Barros JA, Suppes P (2011)
Learning pattern recognition through quasi-synchronization of phase
oscillators.
IEEE Transact on Neural Netw 22:84--95.
doi: 10.1109/TNN.2010.2086476

\item
von der Malsburg C (1981)
The correlation theory of brain function.
Internal Report 81-2, Max-Planck-Institute for Biophysical Chemistry,
G\"ottingen, Germany

\item
Wagemans J, Feldman J, Gepshtein S, Kimchi R, Pomerantz JR.,
van der Helm PA, van Leeuwen C (2012)
A century of Gestalt psychology in visual perception: II. Conceptual and
theoretical foundations.
Psychol Bull 138:1218--1252.
doi: 10.1037/a0029334

\item
Wertheimer, M. (1912).
Experimentelle Studien \"uber das Sehen von Bewegung [On the perception of
motion].
Z f\"ur Psychol 12:161--265

\item
Wertheimer, M. (1923).
Untersuchungen zur Lehre von der Gestalt [On Gestalt theory].
Psychol Forsch 4:301--350

\item
Wolfe JM (2007)
Guided search 4.0: Current progress with a model of visual search.
In: Gray W (ed)
Integrated models of cognitive systems.
Oxford University Press, New York, pp 99--119

\end{list}

\newpage

\section*{Appendix}

Structural information theory (SIT) adopts the simplicity principle, which
holds that the simplest organization of a visual stimulus is the one most
likely perceived by humans.
To enable quantifiable predictions, SIT developed a formal coding model to
determine simplest codes of symbol strings.
This model provides coding rules for the extraction of regularities, and a
metric of the complexity, or structural information load $I$, of codes.

The coding rules serve the extraction of the transparent holographic
regularities repetition (or iteration I), symmetry (S), and alternation (A).
They can be applied to any substring of an input string, and a
code of the input string consists of a string of symbols and coded substrings,
such that decoding the code returns the input string.
Formally, SIT's coding language and complexity metric are defined as follows.

\vspace{\baselineskip}

\small

\noindent
\textbf{Definition 1}
A code $\overline{X}$ of a string $X$ is a string $t_1t_2...t_m$ of code terms
$t_i$ such that $X = D(t_1)...D(t_m)$, where the decoding function
$D : t \rightarrow D(t)$ takes one of the following forms:
\begin{tabbing}
dd \hspace{20mm} \= \hspace{39mm} \= \hspace{45mm} \= \hspace{17mm} \= \kill
\qquad I-form:
\> $n*(\overline{y})$
\> $\rightarrow\quad yyy...y$ \> ($n$ times $y$; $n \geq 2$)
\\[2mm]
\qquad S-form:
\> $S[\overline{(\overline{x_1})(\overline{x_2})...(\overline{x_n})},(\overline{p})]$
\> $\rightarrow\quad x_1x_2...x_n\ p\ x_n...x_2x_1$ \>\> ($n \geq 1$)
\\[2mm]
\qquad A-form:
\> $\langle(\overline{y})\rangle/\langle\overline{(\overline{x_1})(\overline{x_2})...(\overline{x_n})}\rangle$
\> $\rightarrow\quad yx_1\ yx_2\ ...\ yx_n$ \>\> ($n \geq 2$)
\\[2mm]
\qquad A-form:
\> $\langle\overline{(\overline{x_1})(\overline{x_2})...(\overline{x_n})}\rangle/\langle(\overline{y})\rangle$
\> $\rightarrow\quad x_1y\ x_2y\ ...\ x_ny$ \>\> ($n \geq 2$)
\\[2mm]
\qquad Otherwise:
\> $D(t) = t$
\end{tabbing}
for strings $y$, $p$, and $x_i$ ($i = 1,2,...,n$).
The code parts $(\overline{y})$, $(\overline{p})$, and $(\overline{x_i})$ are
\emph{chunks}.
The chunk $(\overline{y})$ in an I-form or an A-form is a \emph{repeat}, and
the chunk $(\overline{p})$ in an S-form is a \emph{pivot} which, as a limit
case, may be empty.
The chunk string $(\overline{x_1})(\overline{x_2})...(\overline{x_n})$ in an
S-form is an \emph{S-argument} consisting of \emph{S-chunks}
$(\overline{x_i})$, and in an A-form, it is an \emph{A-argument} consisting of
\emph{A-chunks} $(\overline{x_i})$.

\vspace{\baselineskip}

\noindent
\textbf{Definition 2}
Let $\overline{X}$ be a code of string $X=s_1 s_2...s_N$.
The {\it complexity} $I$ of $\overline{X}$
in structural information parameters (sip) is given by the sum of (a) the
number of remaining symbols $s_i$ ($1 \leq i \leq N$) and (b) the number
of chunks $(\overline{y})$ in which $y$ is neither a symbol nor an S-chunk.

\normalsize

\vspace{\baselineskip}

The last part of Definition 2 may seem somewhat ad hoc, but has a solid
theoretical basis in terms of degrees of freedom in the hierarchical
organization described by a code.
Furthermore, Definition 1 implies that a string may be encodable into
many different codes.
For instance, a code may involve not only recursive encodings of strings
inside chunks -- that is, from $(y)$ into $(\overline{y})$ -- but also
hierarchically recursive encodings of S- or A-arguments
$(\overline{x_1})(\overline{x_2})...(\overline{x_n})$ into
$\overline{(\overline{x_1})(\overline{x_2})...(\overline{x_n})}$.
The following sample of codes for one and the same string may give a gist
of the abundance of coding possibilities:
\begin{tabbing}
dd \hspace{17mm} \= \hspace{105mm} \= \kill
\qquad String:
\> $X = abacdacdababacdacdab$
\> $I = 20$ sip
\\
\qquad Code 1:
\> $\overline{X} = a\ b\ 2*(acd)\ S[(a)(b),(a)]\ 2*(cda)\ b$
\> $I = 14$ sip
\\
\qquad Code 2:
\> $\overline{X} = \langle(aba)\rangle/\langle(cdacd)(bacdacdab)\rangle$
\> $I = 20$ sip
\\
\qquad Code 3:
\> $\overline{X} = \langle(S[(a),(b)])\rangle/\langle(S[(cd),(a)])(S[(b)(a)(cd),(a)])\rangle$
\> $I = 15$ sip
\\
\qquad Code 4:
\> $\overline{X} = S[(ab)(acd)(acd)(ab)]$
\> $I = 14$ sip
\\
\qquad Code 5:
\> $\overline{X} = S[S[((ab))((acd))]]$
\> $I = 7$ sip
\\
\qquad Code 6:
\> $\overline{X} = 2*(\langle(a)\rangle/\langle S[((b))((cd))]\rangle)$
\> $I = 8$ sip
\end{tabbing}
Code 1 is a code with six code terms, namely, one S-form, two I-forms, and
three symbols.
Code 2 is an A-form with chunks containing strings that may be encoded as
given in Code 3.
Code 4 is an S-form with an empty pivot and illustrates that, in general,
S-forms describe broken symmetry; mirror symmetry then is the limit case in
which every S-chunk contains only one symbol.
Code 5 gives a hierarchical recoding of the S-argument in Code 4.
Code 6 is an I-form in which the repeat has been encoded into an A-form with
an A-argument that has been recoded hierarchically into an S-form.

The computation of a simplest code for a string requires an exhaustive
search for ISA-forms into which its substrings can be encoded, followed by
the selection of a simplest code for the entire string.
This also requires the hierarchical recoding of S- and A-arguments:
A substring of length $k$ can be encoded into $O(2^k)$ S-forms and $O(k2^k)$
A-forms, and to pinpoint a simplest one, simplest codes of the arguments of
these S- and A-forms have to be determined as well -- and so on, with
$O(\log N)$ recursion steps, because $k/2$ is the maximal length of the
argument of an S- or A-form into which a substring of length $k$ can be
encoded.
Hence, recoding S- and A-arguments separately would require
a superexponential $O(2^{N \log N})$ total amount of work.

This combinatorial explosion can be nipped in the bud by gathering the
arguments of S- and A-forms in distributed representations.
The next definitions and proofs show that S-arguments and A-arguments group
naturally into distributed representations consisting of one or more
independent hyperstrings --
which enable the hierarchical recoding of up to an exponential number of
S- or A-arguments in a transparellel fashion, that is, simultaneously as if
only one argument were concerned.
Here, only A-forms $\langle(y)\rangle/\langle(x_1)(x_2)...(x_n)\rangle$
with repeat $y$ consisting of one element are considered, but Definition 4
and Theorem 1 below hold mutatis mutandis for other A-forms as well.

\vspace{4mm}

\begin{list}{}{\setlength{\leftmargin}{4mm}
               \setlength{\labelwidth}{4mm}
               \setlength{\labelsep}{2mm}
               \setlength{\itemsep}{4mm}}
\item[$\bullet$]
Graph-theoretical definition of hyperstrings:

\small

\noindent 
{\bf Definition 3}
A \textit{hyperstring} is a simple semi-Hamiltonian directed acyclic graph
$(V,E)$ with a labeling of the edges in $E$ such that, for all vertices
$i,j,p,q \in V$:
\\
\centerline{either $\pi(i,j) = \pi(p,q)$ or $\pi(i,j) \cap \pi(p,q) = \emptyset$}
\\
where \textit{substring set} $\pi(v_1,v_2)$ is the set of label strings
represented by the paths between vertices $v_1$ and $v_2$; the subgraph
on the vertices and edges in these paths is a \textit{hypersubstring}.

\normalsize

\item[$\bullet$]
Definition of distributed representations called A-graphs, which represent
all A-forms covering suffixes of strings, that is, all A-forms into which
those suffixes can be encoded (see Fig. 9 for an example):

\small

\noindent
{\bf Definition 4}
For a string $T = s_1s_2...s_N$, the \emph{A-graph} $\mathcal{A}(T)$ is a
simple directed acyclic graph $(V,E)$ with $V = \lbrace1,2,..,N+1\rbrace$ and,
for all $1 \leq i < j \leq N$, edges $(i,j)$ and $(j,N+1)$ labeled with,
respectively, the chunks $(s_i...s_{j-1})$ and $(s_j...s_N)$
if and only if $s_i = s_j$.

\normalsize

\item[$\bullet$]
Definition of diafixes, which are substrings centered around the midpoint of
a string (this notion complements the known notions of prefixes and suffixes,
and facilitates the explication of the subsequent definition of S-graphs):

\small

\noindent
{\bf Definition 5}
A \emph{diafix} of a string $T = s_1s_2...s_N$ is a substring
$s_{i+1}...s_{N-i}$ ($0 \leq i < N/2$).

\normalsize

\item[$\bullet$]
Definition of distributed representations called S-graphs, which represent
all S-forms covering diafixes of strings (see Fig. 10 for an example):

\small

\noindent
{\bf Definition 6}
For a string $T = s_1s_2...s_N$, the \emph{S-graph} $\mathcal{S}(T)$ is a
simple directed acyclic graph $(V,E)$ with
$V = \lbrace1,2,..,\lfloor N/2 \rfloor + 2\rbrace$ and, for all
$1 \leq i < j < \lfloor N/2 \rfloor + 2$, edges $(i,j)$ and
$(j,\lfloor N/2 \rfloor + 2)$ labeled with, respectively, the chunk
$(s_i...s_{j-1})$ and the possibly empty chunk $(s_j...s_{N-j+1})$
if and only if $s_i...s_{j-1} = s_{N-j+2}...s_{N-i+1}$.
\end{list}

\normalsize

\begin{figure}

\small

\begin{center}
\includegraphics[width=14cm]{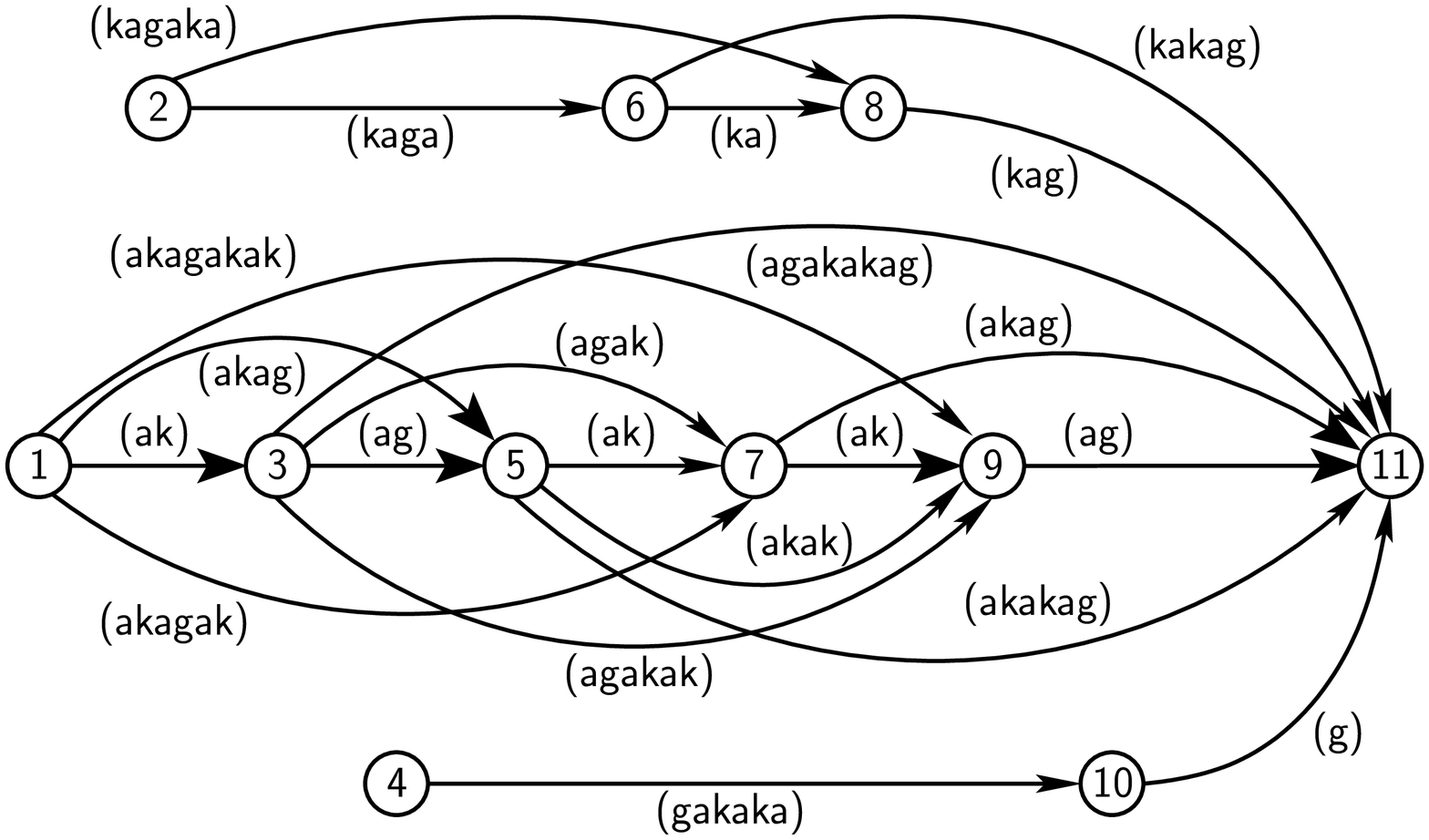}
\end{center}

\textbf{Fig. 9}
The A-graph for string $T = akagakakag$, with three independent hyperstrings
(connected only at vertex $11$) for the three sets of A-forms with repeats
$a$, $k$, and $g$, respectively, which cover suffixes of $T$.
An A-graph may contain so-called pseudo A-pair edges -- like, here, edge
$(10,11)$ -- which do not correspond to actual repeat plus A-chunk pairs;
they cannot end up in codes but are needed to maintain the
integrity of hyperstrings during recoding

\vspace{\baselineskip}
\vspace{\baselineskip}

\begin{center}
\includegraphics[width=11cm]{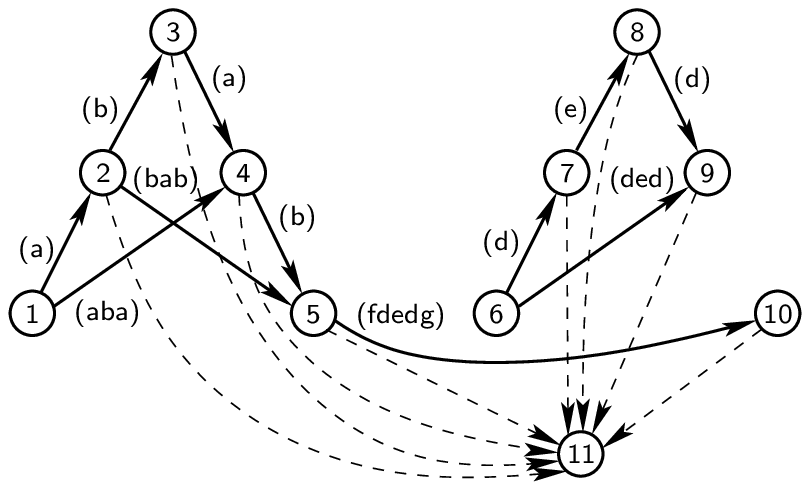}
\end{center}

\textbf{Fig. 10}
The S-graph for string $T=\textit{ababfdedgpfdedgbaba}$, with two independent
hyperstrings given by the solid edges, which represent S-chunks in S-forms
covering diafixes of $T$.
The dashed edges represent the pivots, which come into play after
hyperstring recoding

\normalsize

\end{figure}

\noindent
\textbf{Theorem 1}
The A-graph $\mathcal{A}(T)$ for a string $T = s_1s_2...s_N$ consists of at
most $N+1$ disconnected vertices and at most $\lfloor N/2 \rfloor$
independent subgraphs (i.e., subgraphs that share only the sink vertex
$N+1$), each of which is a hyperstring.
\\[4mm]
\textit{Proof}
(1)
By Definition 4, vertex $i$ ($i \leq N$) in $\mathcal{A}(T)$
does not have incoming or outgoing edges if and only if $s_i$ is a unique
element in $T$.
Since $T$ contains at most $N$ unique elements, $\mathcal{A}(T)$ contains
at most $N+1$ disconnected vertices, as required.

(2)
Let $s_{i_1},s_{i_2},...,s_{i_n}$ ($i_p < i_{p+1}$) be a complete
set of identical elements in $T$.
Then, by Definition 4, the vertices $i_1,i_2,...,i_n$ in
$\mathcal{A}(T)$ are connected with each other and with vertex $N+1$ but not
with any other vertex.
Hence, the subgraph on the vertices $i_1,i_2,...,i_n,N+1$ forms an
independent subgraph.
For every complete set of identical elements in $T$, $n$ may be as small as
$2$, so that $\mathcal{A}(T)$ contains at most $\lfloor N/2 \rfloor$
independent subgraphs, as required.

(3)
The independent subgraphs must be semi-Hamiltonian to be hyperstrings.
Now, let $s_{i_1},s_{i_2},...,s_{i_n}$ ($i_p < i_{p+1}$) again be a
complete set of identical elements in $T$.
Then, by Definition 4, $\mathcal{A}(T)$ contains edges
$(i_p,i_{p+1})$, $p = 1,2,...,n-1$, and it contains edge $(i_n,N+1)$.
Together, these edges form a Hamiltonian path through the independent subgraph
on the vertices $i_1,i_2,...,i_n,N+1$, as required.

(4)
The only thing left to prove is that the substring sets are pairwise
either identical or disjoint.
Now, for $i < j$ and $k \geq 1$, let substring sets $\pi(i,i+k)$ and
$\pi(j,j+k)$ in $\mathcal{A}(T)$ be not disjoint, that is, let them share
at least one chunk string.
Then, the substrings $s_{i}...s_{i+k-1}$ and $s_{j}...s_{j+k-1}$ of $T$ are
necessarily identical and, also necessarily, $s_{i} = s_{i+k}$ and either
$s_{j} = s_{j+k}$ or $j+k = N+1$.
Hence, by Definition 4, these identical substrings of $T$ yield, in
$\mathcal{A}(T)$, edges $(i,i+k)$ and $(j,j+k)$ labeled with the identical
chunks $(s_{i}...s_{i+k-1})$ and $(s_{j}...s_{j+k-1})$, respectively.
Furthermore, obviously, these identical substrings of $T$ can be chunked into
exactly the same strings of two or more identically beginning chunks.
By Definition 4, all these chunks are represented in $\mathcal{A}(T)$,
so that each of these chunkings is represented not only by a path $(i,...,i+k)$
but also by a path $(j,...,j+k)$.
This implies that the substring sets $\pi(i,i+k)$ and $\pi(j,j+k)$ are
identical.
The foregoing holds not only for the entire A-graph but, because of their
independence, also for every independent subgraph.
Hence, in sum, every independent subgraph is a hyperstring, as required.
$\blacksquare$

\newpage

\noindent
\textbf{Lemma 1}
Let the strings $c_1 = s_1s_2...s_k$ and $c_2 = s_1s_2...s_p$ ($k<p$) be such
that $c_2$ can be written in the following two ways:

\begin{center}
$c_2 = c_1X$ \quad with \quad $X = s_{k+1}...s_p$
\\
$c_2 = Yc_1$ \quad with \quad $Y = s_1...s_{p-k}$
\end{center}

\noindent
Then, $X=Y$ if $q=p/(p-k)$ is an integer; otherwise $Y=VW$ and $X=WV$, where
$V = s_1...s_r$ and $W = s_{r+1}...s_{p-k}$,
with $r= p- \lfloor q \rfloor (p-k)$.
\\[4mm]
\textit{Proof}
(1) If $1<q<2$, then $c_2=c_1Wc_1$, so that $Y=c_1W$ and $X=Wc_1$.
Then, too, $r=k$, so that $c_1=V$.
Hence, $Y=VW$ and $X=WV$, as required.

(2) If $q=2$, then $c_2=c_1c_1$.
Hence, $X=Y=c_1$, as required.

(3) If $q>2$, then the two copies of $c_1$ in $c_2$ overlap each
other as follows:

\vspace{\baselineskip}

\begin{math}
\begin{array}{lccccccccccc}
c_2 = c_1 X =
s_1  & \dots & s_{p-k} & s_{p-k+1} & \dots & s_k      &  s_{k+1}   & \dots & s_p
\\
c_2 = Y c_1 =
    & Y     &         & s_1       & \dots & s_{2k-p} & s_{2k-p+1} & \dots & s_k
\end{array}
\end{math}

\vspace{\baselineskip}

\noindent
Hence, $s_i=s_{p-k+i}$ for $i=1,2,...,k$.
That is, $c_2$ is a prefix of an infinite repetition of $Y$.

(3a) If $q$ is an integer, then $c_2$ is a $q$-fold repetition of
$Y$, that is, $c_2=YY...Y$.
This implies (because also $c_2=Yc_1$) that $c_1$ is a $(q-1)$-fold repetition
of $Y$, so, $c_2$ can also be written as $c_2=c_1Y$.
This implies $X=Y$, as required.

(3b) If $q$ is not an integer, then $c_2$ is a
$\lfloor q \rfloor$-fold repetition of $Y$ plus a residual prefix $V$ of $Y$,
that is, $c_2=YY...YV$.
Now, $Y=VW$, so that $c_2$ can also be written as $c_2=VWVW...VWV$.
This implies (because also $c_2=Yc_1=VWc_1$) that $c_1=VW...VWV$, that is,
$c_1$ is a $(\lfloor q \rfloor - 1)$-fold repetition of $Y = VW$ plus a
residual part $V$.
This, in turn, implies that $c_2$ can also be written as $c_2=c_1WV$, so that
$X=WV$, as required.
$\blacksquare$

\newpage

\noindent
\textbf{Lemma 2}
Let $\mathcal{S}(T)$ be the S-graph for a string $T = s_1s_2...s_N$.
Then:
\\
(1) If $\mathcal{S}(T)$ contains edges $(i,i+k)$ and $(i,i+p)$,
with $k < p < \lfloor N/2 \rfloor + 2 - i$, then it also
contains a path $(i+k,...,i+p)$.
\\
(2) If $\mathcal{S}(T)$ contains edges $(i-k,i)$ and $(i-p,i)$, with $k<p$ and
$i < \lfloor N/2 \rfloor + 2$, then it also contains a path $(i-p,...,i-k)$.
\\[4mm]
\textit{Proof}
(1)
Edge $(i,i+k)$ represents S-chunk $(c_1) = (s_i...s_{i+k-1})$, and edge
$(i,i+p)$ represents S-chunk $(c_2) = (s_i...s_{i+p-1})$.
This implies that diafix $D = s_i...s_{N-i+1}$ of $T$ can be written in
two ways:
\begin{center}
\begin{math}
\begin{array}{ccccc}
D & = & c_2 & \dots & c_2
\\
D & = & c_1 & \dots & c_1
\end{array}
\end{math}
\end{center}
This implies that $c_2$ (which is longer than $c_1$) can be written
in two ways:
\begin{center}
$c_2 = c_1X$ \quad with \quad $X = s_{i+k}...s_{i+p-1}$
\\
$c_2 = Yc_1$ \quad with \quad $Y = s_i...s_{i+p-k-1}$
\end{center}
Hence, by Lemma 1, either $X = Y$ or $Y = VW$ and $X =WV$ for some $V$
and $W$.
If $X = Y$, then $D = c_1Y...Yc_1$ so that, by Definition 6, $Y$ is an
S-chunk represented by an edge that yields a path $(i+k,...,i+p)$ as required.
If $Y = VW$ and $X =WV$, then $D=c_1WV...VWc_1$ so that, by
Definition 6, $W$ and $V$ are S-chunks represented by
subsequent edges that yield a path $(i+k,...,i+p)$ as required.

(2)
This time, edge $(i-k,i)$ represents S-chunk
$(c_1) = (s_{i-k}...s_{i-1})$, and edge $(i-p,i)$ represents S-chunk
$(c_2) = (s_{i-p}...s_{i-1})$.
This implies that diafix $D = s_{i-p}...s_{N-i+p+1}$ of $T$ can be written
in two ways:
\begin{center}
\begin{math}
\begin{array}{ccccc}
D & = & c_2 & \dots & c_2
\\
D & = & Yc_1 & \dots & c_1X
\end{array}
\end{math}
\end{center}
with $X$=$s_{i-p+k}...s_{i-1}$ and $Y$=$s_{i-p}...s_{i-k-1}$.
Hence, as before, $c_2 = c_1X$ and $c_2 = Yc_1$, so that, by Lemma 1,
either $X=Y$ or $Y = VW$ and $X =WV$ for some $V$ and $W$.
This implies either $D=Yc_1...c_1Y$ or $D=VWc_1...c_1WV$.
Hence, this time, Definition 6 implies that both cases yield a path
$(i-p,...,i-k)$, as required.
$\blacksquare$

\newpage

\noindent
\textbf{Lemma 3}
In the S-graph $\mathcal{S}(T)$ for a string $T = s_1s_2...s_N$, the substring
sets $\pi(v_1,v_2)$ ($1 \leq v_1 < v_2 < \lfloor N/2 \rfloor + 2$) are
pairwise identical or disjoint.
\\[4mm]
\textit{Proof}
Let, for $i < j$ and $k \geq 1$, substring sets $\pi(i,i+k)$ and $\pi(j,j+k)$
in $\mathcal{S}(T)$ be nondisjoint, that is, let them share at least one
S-chunk string.
Then, the substrings $s_{i}...s_{i+k-1}$ and $s_{j}...s_{j+k-1}$ in the
left-hand half of $T$ are necessarily identical to each other.
Furthermore, by Definition 6, the substring in each chunk of these
S-chunk strings is identical to its symmetrically positioned counterpart in the
right-hand half of $T$, so that also the substrings $s_{N-i-k+2}...s_{N-i+1}$
and $s_{N-j-k+2}...s_{N-j+1}$ in the right-hand half of $T$ are identical to
each other.
Hence, the diafixes $D_1=s_{i}...s_{N-i+1}$ and $D_2=s_{j}...s_{N-j+1}$ can be
written as
\begin{center}
$D_1 = s_{i}...s_{i+k-1}\;\;p_1\;\;s_{N-i-k+2}...s_{N-i+1}$
\\
$D_2 = s_{i}...s_{i+k-1}\;\;p_2\;\;s_{N-i-k+2}...s_{N-i+1}$
\end{center}
with $p_1=s_{i+k}...s_{N-i-k+1}$ and $p_2=s_{j+k}...s_{N-j-k+1}$.
Now, by means of any S-chunk string $C$ in $\pi(i,i+k)$, diafix $D_1$ can be
encoded into the covering S-form $S[C,(p_1)]$.
If pivot $(p_1)$ is replaced by $(p_2)$, then one gets
the covering S-form $S[C,(p_2)]$ for diafix $D_2$.
This implies that any S-chunk string in $\pi(i,i+k)$ is in $\pi(j,j+k)$,
and vice versa.
Hence, nondisjoint substring sets $\pi(i,i+k)$ and $\pi(j,j+k)$ are identical,
as required.
$\blacksquare$

\vspace{\baselineskip}

\noindent
\textbf{Theorem 2}
The S-graph $\mathcal{S}(T)$ for a string $T = s_1s_2...s_N$
consists of at most $\lfloor N/2 \rfloor + 2$ disconnected vertices and
at most $\lfloor N/4 \rfloor$ independent subgraphs that, without the sink
vertex $\lfloor N/2 \rfloor + 2$ and its incoming pivot edges, form one
disconnected hyperstring each.
\\[4mm]
\textit{Proof}
From Definition 6, it is obvious that there may be disconnected
vertices and that their number is at most $\lfloor N/2 \rfloor + 2$, so
let us turn to the more interesting part.
If $\mathcal{S}(T)$ contains one or more paths $(i,...,j)$
($i<j<\lfloor N/2 \rfloor+2$) then, by Lemma 2, one of these paths
visits every vertex $v$ with $i<v<j$ and $v$ connected to $i$ or $j$.
This implies that, without the pivot edges and apart from disconnected
vertices, $\mathcal{S}(T)$ consists of disconnected semi-Hamiltonian subgraphs.
Obviously, the number of such subgraphs is at most $\lfloor N/4 \rfloor$, and
if these subgraphs are expanded to include the pivot edges, they form one
independent subgraph each.
More important, by Lemma 3, these disconnected semi-Hamiltonian subgraphs
form one hyperstring each, as required.
$\blacksquare$

\end{document}